\documentclass[10pt,journal,compsoc]{IEEEtran} 
\usepackage{amsmath,amsfonts}
\usepackage{algorithmic}
\usepackage{algorithm}
\usepackage{array}

\usepackage{textcomp}
\usepackage{stfloats}
\usepackage{url}
\usepackage{verbatim}
\usepackage{graphicx}
\usepackage{cite}

\usepackage{url}
\usepackage{booktabs} 
\usepackage{amsfonts}  
\usepackage{nicefrac}   
\usepackage{microtype}    
\usepackage{xcolor}   
\usepackage{graphicx}
\usepackage{enumitem}

\definecolor{myblue}{rgb}{0.21,0.49,0.74}
\definecolor{myred}{rgb}{0.54, 0.04, 0.21}
\definecolor{myorange}{rgb}{0.87, 0.51, 0.26}
\usepackage[pagebackref,breaklinks,colorlinks,citecolor=myblue]{hyperref}

\usepackage{xspace}
\usepackage{xfrac}
\usepackage{multirow}
\usepackage{makecell}
\usepackage{pifont}
\usepackage{colortbl}
\usepackage{amsmath}
\usepackage{wrapfig}
\usepackage[capitalise]{cleveref}
\usepackage{etoc}
\usepackage{titletoc}

\usepackage{multirow}
\usepackage[table]{xcolor}
\usepackage{algorithm} 
\usepackage{amssymb}
\usepackage{amsfonts}
\usepackage{multicol}
\usepackage{graphicx}
\usepackage{booktabs}
\usepackage[T1]{fontenc}

\usepackage[numbers,sort]{natbib}

\usepackage{comment}

\usepackage{subcaption}
\usepackage{amsmath}
\usepackage{algorithmic}
\usepackage{array}
\usepackage[switch]{lineno}
\usepackage{longtable}
\usepackage{xspace}
\usepackage{url}
\usepackage{overpic}
\usepackage{ragged2e}
\usepackage{framed}
\usepackage{enumitem}

\usepackage[normalem]{ulem} 
\newcommand{\mrev}[1]
{{#1}}

\makeatletter
\DeclareRobustCommand\onedot{\futurelet\@let@token\@onedot}
\def\@onedot{\ifx\@let@token.\else.\null\fi\xspace}

\def\eg{\emph{e.g}\onedot} 

\def\ie{\emph{i.e}\onedot}

\makeatother

\definecolor{LightCyan}{rgb}{0.87,0.92,0.96}
\definecolor{lightgray}{gray}{.9}
\newcommand{\tablestyle}[2]{\setlength{\tabcolsep}{#1}\renewcommand{\arraystretch}{#2}\centering\footnotesize}
\usepackage{subcaption}
\usepackage{lipsum}
\usepackage{bbding}
\usepackage{diagbox}
\newcommand{\algname}{TTC\xspace}

\newcommand{\modelnamefull}{Online Adapter\xspace}
\newcommand{\modelname}{OA module\xspace}
\newcommand{\myparagraph}[1]{\vspace{1pt}\noindent{\bf #1}}

\begin{document}

\renewcommand{\thetable}{\arabic{table}}
\renewcommand{\thefigure}{\arabic{figure}}

\setcounter{figure}{0}
\setcounter{table}{0}

\twocolumn

\title{Test-time Correction: 
An Online \\ 3D Detection System 
via Visual Prompting}

\author{
    Hanxue Zhang$^{*}$,
    Zetong Yang$^{*}$,
    Yanan Sun,
    Li Chen,
    Fatma G\"uney
    and Hongyang Li%
    \thanks{$^{*}$ Equal contribution.}%
    \IEEEcompsocitemizethanks{
        \IEEEcompsocthanksitem 
        Hanxue Zhang is with Shanghai Jiao Tong University, Shanghai, China, 
        and also with Shanghai AI Lab, Shanghai, China. 
        E-mail: jjxjiaxue@gmail.com.

        \IEEEcompsocthanksitem 
        Zetong Yang is with GAC R\&D Center, Guangzhou, China. 
        E-mail: tomztyang@gmail.com.

        \IEEEcompsocthanksitem 
        Yanan Sun is with Shanghai AI Lab, Shanghai, China. 
        E-mail: now.syn@gmail.com.        

        \IEEEcompsocthanksitem 
        Li Chen and Hongyang Li are with The University of Hong Kong, Hong Kong, China. 
        E-mail: hongyang@hku.hk, ilnehc@connect.hku.hk.   

        \IEEEcompsocthanksitem 
        Fatma G\"uney is with Ko\c{c} University, Turkey. 
        E-mail: fguney@ku.edu.tr.

        \IEEEcompsocthanksitem 
        Primary contact: \texttt{hongyang@hku.hk}
    }
}

\IEEEtitleabstractindextext{%
\begin{abstract}
This paper introduces Test-time Correction (\algname), an online 3D detection system designed to rectify test-time errors using various auxiliary feedback, aiming to enhance the safety of deployed autonomous driving systems.
Unlike conventional offline 3D detectors that remain fixed during inference, \algname enables immediate online error correction without retraining, allowing autonomous vehicles to adapt to new scenarios and reduce deployment risks.
To achieve this, we equip existing 3D detectors with an Online Adapter (OA) module—a prompt-driven query generator for real-time correction.
At the core of \modelname are visual prompts: image-based descriptions of objects of interest derived from auxiliary feedback such as mismatches with 2D detections, road descriptions, or user clicks.
These visual prompts, collected from risky objects during inference, are maintained in a visual prompt buffer to enable continuous correction in future frames.
By leveraging this mechanism, \algname consistently detects risky objects, achieving reliable, adaptive, and versatile driving autonomy.
Extensive experiments show that \algname significantly improves instant error rectification over frozen 3D detectors, even under limited labels, zero-shot settings, and adverse conditions.
We hope this work inspires future research on post-deployment online rectification systems for autonomous driving.
\end{abstract}

}

\maketitle

\IEEEpeerreviewmaketitle

\IEEEraisesectionheading{
\section{Introduction}\label{sec:introduction}
}

\IEEEPARstart{V}{isual-based} 3D object detection, which localizes and classifies 3D objects from visual imagery, plays a crucial role in autonomous driving systems. 
Visual autonomous driving frameworks \cite{hu2023_uniad,hu2022stp3,casas2021mp3,Cui_2021} rely heavily on accurate 3D detection outcomes to predict future driving behaviors and plan the trajectory of the ego vehicle.
Existing 3D object detectors \cite{li2022bevformer,yang2023bevformerV2,liu2022petr,CaDDN,huang2021bevdet} typically follow an offline training and deployment pipeline. 
Once the model is trained and deployed on self-driving cars, it is expensive to update new behaviors, \eg, another turn of offline re-training or fine-tuning. 
That is, when the system fails to perceive important objects or fails in novel scenarios due to a domain shift, these offline solutions cannot update themselves online to rectify mistakes immediately and detect missed objects.
Such a caveat poses significant safety risks to reliable driving systems, \eg, dangerous driving behaviors such as improper lane changing, turning, or even collisions.

\begin{figure}[htbp]
 \centering
    \includegraphics[width=0.48\textwidth]{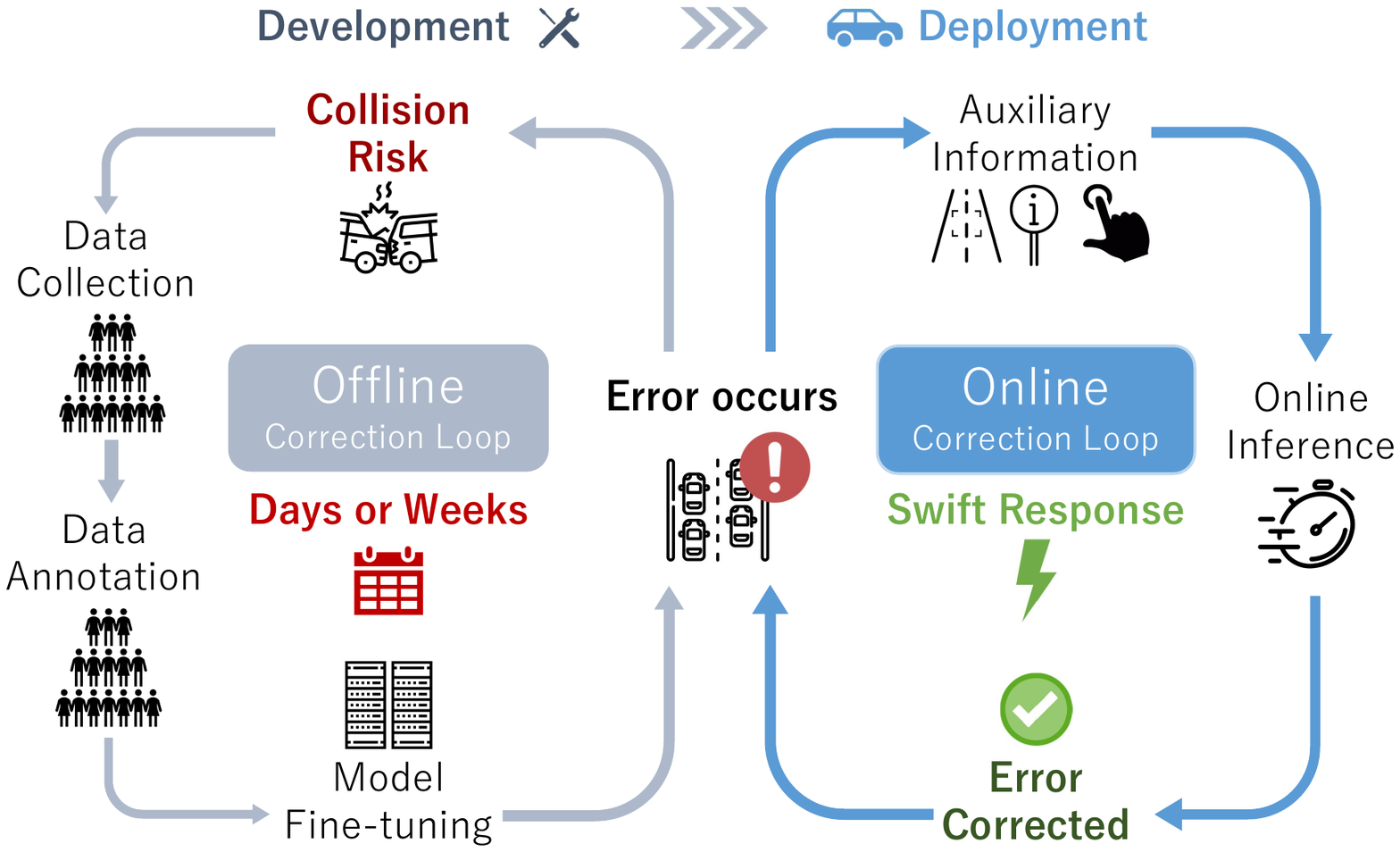}
  \caption{\textbf{Comparison of Error Correction} between the conventional offline loop (left) and the new proposed 
  online \algname System (right). 
  Offline error correction pipeline improves model capability during the development stage, which typically requires expensive workloads and computational overhead over days or weeks for model updates.
  While \algname system additionally enables deployed 3D detectors with on-the-fly error rectification ability.}
  \label{fig:teaser}
\end{figure}

To guarantee safety, we argue that 3D detectors deployed on autonomous driving systems can rectify missed detection on the fly during test time.
As depicted in \Cref{fig:teaser} (left), for error rectification, existing 3D detectors rely on offline pipelines, encompassing a full-suite procedure of data collection, annotation, training, and deployment. 
This requires significant human workloads and resources for labeling and re-training, days or even weeks to fulfill.
Other than the offline pipeline to improve models, we desire deployed 3D detectors also capable of test-time correction since such delays in offline updating are unacceptable when facing risks on the road, where safety is of the utmost priority.

In this work, we propose a novel 3D detection system, \algname, that provides online \textbf{T}est-\textbf{T}ime \textbf{C}orrection by leveraging various auxiliary information (\eg, 2D detections, road descriptions, or human feedback), as shown in~\Cref{fig:teaser} (right). 
The system is designed to address missed detection issues in autonomous driving scenarios, enabling existing 3D detectors to leverage various real-time feedback from multiple sources for instant error correction, thereby reducing the likelihood of erroneous decisions caused by undetected objects.
Inspired by the principles of In-context Learning Customization \cite{cot,llm2023peng} in large language models (LLMs) \cite{gpt4,gemini}, 
we implement test-time correction by converting auxiliary information into prompts and feeding them into the system for online adaptation.

\begin{figure}[t]
 \centering
  \vspace{-0.in}
    \includegraphics[width=0.38\textwidth]{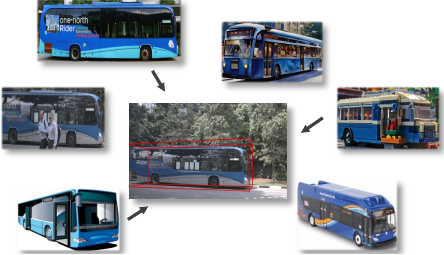}
  \caption{\textbf{Visual prompts} could be arbitrary views of objects, across zones, styles, timestamps, \textit{etc}.}
  \label{fig:teaser_visual_prompt}
\end{figure}

The proposed \algname includes two components: \modelnamefull (OA) that enables 3D detectors with visual promotable ability, and a visual prompt buffer that records risky objects.
The core design is ``visual prompts'', the visual object representation derived from various auxiliary sources. 
Existing promptable 3D detection methods typically utilize text, boxes, or clicks as prompts \cite{chen2024ll3da,choi2024idet3d,zhang2023sam3d,guo2024sam2point,cao2024coda}. 
However, text prompts often require integration with large language models, and supplying a text prompt on a per-frame basis results in significant latency.
Meanwhile, box and point prompts on specific frames struggle to handle streaming data.
These limitations indicate that such prompts are inadequate for real-time autonomous driving tasks.
Visual prompts cover arbitrary imagery views of target objects, \ie, views in different zones, styles, and timestamps (\Cref{fig:teaser_visual_prompt}), indicating the identity of target objects.
Once generated, visual prompts can operate on subsequent frames, accurately describing and tracking target risky objects.
With visual prompts, \modelname generates corresponding queries, locates corresponding objects within streaming inputs, and facilitates 3D detectors to output 3D boxes.

To enable consecutive error rectification for video streaming, we design a dynamic visual prompt buffer to maintain visual prompts of all risky objects.
In each iteration, we use all visual prompts in this buffer as inputs to the \algname system. 
This enables the continuous detection and tracking of all risky objects, redressing online errors for streaming input effectively.
Further, to prevent the buffer from undesirable expansion, we introduce a ``dequeue'' operation to ensure its bounded size, allowing for consecutive rectification without excessive overhead.

We conduct experiments on the nuScenes dataset \cite{nuscenes2019}.
Given the novel setting of our \algname system, the assessment focuses on its abilities for instant online error correction. 
To this end, we design extensive experiments to verify the key aspects of the overall system: the effectiveness of \algname system in rectifying errors over the online video stream; and the effectiveness of \algname when encountering challenging, extreme scenarios with large amounts of missed detections.

Remarkably, the \algname system significantly improves the offline 3D detectors for test-time performance. 
Specifically, with test-time rectification, \algname improves offline monocular \cite{zhang2022monodetr}, multi-view \cite{wang2023object}, and BEV detectors \cite{yang2023bevformerV2} by 2.4\%, 9.7\%, and 9.3\% mAP, respectively.
Second, on challenging scenarios, the \algname system exhibits even more substantial gains, with improvements of 10.1\%, 20.0\%, 7.2\%, and 3.8\% mAP on tasks like distant 3D detection and vehicle-focused detection with limited annotations, zero-shot extensions, as well as scenarios with domain shifts, respectively.

This comprehensive evaluation highlights the versatility and adaptability of \algname, which can effectively rectify online errors and maintain robust 3D detection even under limited data, category shifts, and environmental changes.
We hope the introduction of \algname will inspire the community to further explore the online rectification approach in autonomous driving systems, a crucial technology that can enhance the safety and reliability of safety-critical applications.

\section{Related Work}

In this paper, we introduce the task of online 3D detection, which focuses on the instant rectification of errors during the online testing phase of 3D detectors. The primary goal is to enable the continuous detection of objects missed by offline-trained 3D detectors, without requiring additional training after deployment. This work is closely related to several areas, including 3D detection, tracking, test-time adaptation, and interactive vision models.

\subsection{3D Object Detection}
3D object detection is a cornerstone task in autonomous driving. 
Traditional 3D detectors primarily focus on the performance of pre-trained models \cite{PVRCNN,lang2018pointpillars,yin2021center,philion2020lift,li2022bevformer}. 
Once deployed, the model's output is fixed during inference, with no ability to intervene in real time to correct missed detections.
In recent years, some works have explored instruction-based detection systems, offering interfaces that accept prompts \cite{chen2024ll3da,choi2024idet3d,zhang2023sam3d,guo2024sam2point,zhu2025unifying,cao2024coda,wang2024omnidrive}. However, these systems typically rely on boxes, points, or language as interfaces. Boxes and points struggle to handle temporal data \cite{choi2024idet3d,zhang2023sam3d,guo2024sam2point}, while language prompts often depend on LLMs \cite{chen2024ll3da,zhu2025unifying,cao2024coda,wang2024omnidrive}, leading to higher latency and potential ambiguities. 
In contrast, online 3D detection focuses on real-time error correction during sequential detection.
By utilizing visual prompts, we enable accurate correction of target objects in streaming data instantly.

\subsection{Target Object Tracking}
Tracking is another direction closely related to online 3D detection \cite{pnpnet,pang2021simpletrack,Multiple3Dtracking,zhang2022mutr3d,Hu2021QD3DT}. 
Since our approach uses visual prompts to identify targets, it closely resembles single object tracking (SOT) in terms of its objectives \cite{yan2021learning,chen2023seqtrack,cui2022mixformer,cui2024mixformerv2}.
While current SOT methods based on images are limited to generating 2D detection outputs and presume the object query as the initial frame of the ongoing video, our \algname system stands out. It can directly identify and track the associated 3D bounding boxes using visual prompts, even when these prompts originate from varied sources, scenes, and zones with distinct styles.

\subsection{Interactive Vision Models}

Interactive vision models are designed for tasks based on user inputs.
As one of the fundamental tasks in computer vision, they are extensively researched with numerous breakthroughs~\cite{lazysnap,focalclick,simpleclick,deepinteract,randomwalk}. 
Particularly, the advent of the Segment Anything Model (SAM) \cite{kirillov2023segany} has sparked a surge of progress, with applications spanning reconstruction \cite{shen2023anything3d}, detection \cite{ren2024grounded,yang2023track}, segmentation~\cite{zou2023segment}, image editing \cite{gao2023editanything}, and more.
Compared to existing models, which typically rely on prompts such as clicks, boxes, or scribbles, in this work, we study visual prompts, a new prompt referring to the actual images of objects with arbitrary poses and styles.
Visual prompts enable continuous detection and tracking of target objects in streaming videos, facilitating the immediate correction of failed detections at test time.
This represents a departure from traditional prompt types, offering more natural and dynamic interactions with the visual content.

\subsection{Test-time Adaptation}
\label{sec:re-tta}

Test-time adaptation aims to improve model performance on test data by adapting the model to test samples, even in the presence of data shifts \cite{zhang2023adanpc,niu2022efficient,niutowards,schneider2020improving,wang2020tent,liu2021ttt++}.
\mrev{Most TTA methods update network parameters during testing—via entropy minimization, self-supervised objectives, or batch-norm statistic adaptation. 
More recently, several works have brought TTA to object detection \cite{ruan2024fully,vs2023towards,cao2024exploring}, and to 3D detection in particular (e.g., MonoTTA \cite{lin2025monotta}, Reg-TTA3D~\cite{yuan2024reg}, and MOS \cite{chenmos}). }
\mrev{These approaches typically fine-tune on streaming target batches across domains or perform input-conditioned model selection/mixture, still relying on weight updates or model switching at test time. }
\mrev{In contrast, our method does not modify the original model parameters. Instead, it 
performs parameter-free, prediction-level rectification driven by various auxiliary feedback, enabling instance-level online correction without retraining or compromising source-domain accuracy.}


\section{Test-time Correction}

\begin{figure*}[t]
	\centering
        \includegraphics[width=1.0\linewidth]{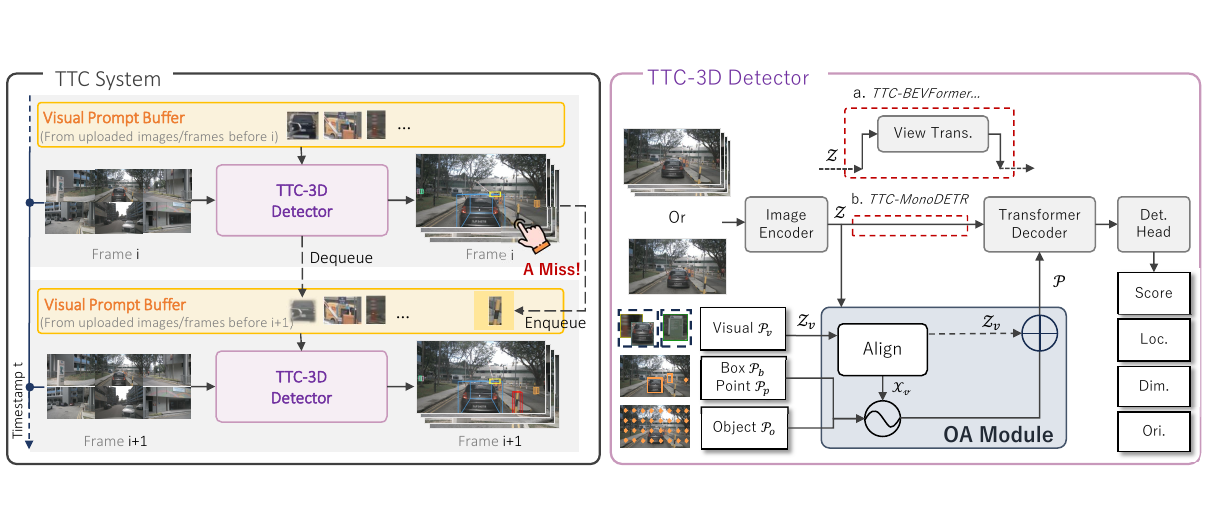}
	\vspace{-0.15in}
	\caption{\textbf{Overall Framework.} (Left:) The \algname system centers on a TTC-3D Detector which utilizes visual prompts $\mathcal{P}_v$ from the visual prompt buffer for test-time error rectification.
 (Right:) The TTC-3D Detector can be based on any traditional detector (BEV or monocular).
 It supports 3D detection from any combination of four prompts, \ie, object $\mathcal{P}_o$, box $\mathcal{P}_b$, point $\mathcal{P}_p$, and novel visual prompts $\mathcal{P}_v$, arbitrary views of target objects across scenarios and timestamps.}
	\vspace{-0.15in}
	\label{fig:framework}
\end{figure*}

In this section, we elaborate on our \algname, an online test-time error rectification system for 3D detection to detect and track missed objects during on-road inference with the guidance of various feedback. 
We start with an overview in \Cref{sec:sec3.1}, then delve into \modelname and the visual prompt buffer in \Cref{sec:sec3.2}, and \Cref{sec:sec3.3}, respectively.

\subsection{Overview}
\label{sec:sec3.1}

We convert various online auxiliary feedback, \ie, such as human-provided clicks and boxes, road descriptions, and misaligned detection results compared with 2D detectors, into a uniform representation called ``visual prompts'', image descriptions of missed or risky objects. 
Such image-based descriptions can cover arbitrary views of objects, including pictures taken from diverse zones, weather, timestamps, or even from out-of-domain sources such as stylized Internet images.
Based on visual prompts, the \algname system is designed as a real-time framework capable of continuously detecting and tracking risky objects, even when the raw feedback may suffer from high latency.
As \Cref{fig:framework} (left) shows, it comprises two key components: \textit{(i)} \algname-3D Detector, any 3D detector equipped with \modelname for in-context 3D detection and tracking via visual prompts, and \textit{(ii)} an extendable visual prompt buffer storing visual prompts of all previously missed objects, enabling continuous online error rectification.

During online inference after deployment in vehicles, the model can leverage auxiliary information by converting it into visual prompts and adding them to the visual prompt buffer.
For user clicks, the model first detects the corresponding 2D boxes based on the user-provided point prompt and updates the visual prompt buffer with the associated image patch.
For user-drawn boxes or cases involving misaligned 2D detection results, the corresponding image regions can be directly pre-processed via cropping.
For road descriptions, relevant visual prompts can be retrieved via online search based on the provided textual descriptions.
\algname-3D Detector leverages the stored visual prompts to detect and track risky 3D objects.
This enables instant error correction and continuous improvement of 3D detection during online operation. 

\subsection{\modelnamefull (OA)}
\label{sec:sec3.2}

\modelname is designed as a plug-and-play module.
It receives prompts and transforms them into queries that can be processed by traditional detectors.
When trained with \modelname, traditional detectors can understand various prompts to produce detection results, enabling online error correction without the need for parameter updates.

\myparagraph{Prompt-driven Query Generating.}
As shown in \Cref{fig:framework} (right), \modelname is flexible to handle four prompts in different forms: object query prompts for traditional offline 3D detection, box and point prompts for feedback converting, and visual prompts for consistent error correction. 
Specifically, these prompts are processed as follows:

\begin{itemize} [itemsep=0pt]
    \item For object prompts $\mathcal{P}_o$, the \modelname generates a set of learnable embeddings as queries, akin to traditional 3D detectors, which are updated during training as demonstrated in previous works \cite{li2022bevformer, detr3d, zhang2022monodetr}.
    \item For box $\mathcal{P}_b$ and point $\mathcal{P}_p$ prompts, the \modelname encodes them with their location and shape, representing them as Fourier features \cite{tancik2020fourfeat}.
    \item For visual prompts $\mathcal{P}_v$, the \modelname first extracts the visual prompt features $\mathcal{Z}_v$ by an encoder \cite{ResNet}, then localizes their corresponding objects within the input images $\mathcal{X}_v$, and finally adds their features with the Fourier positional encoding as subsequent inputs:
\begin{equation} \label{eq:visualposenc}
	\begin{aligned}
            \mathcal{P}_v = \texttt{FourierPE}(\texttt{Align}(\mathcal{Z}_v, \mathcal{Z})) + \mathcal{Z}_v,
	\end{aligned}
\end{equation} 
where $\mathcal{Z}$ means image features of image input, $\texttt{FourierPE}(\cdot)$ is the Fourier positional encoding and the $\texttt{Align}(\cdot)$ operation means ``Visual Prompt Alignment'', the process of inferring the 2D position in the current frame of the visual prompt.
\end{itemize}

\myparagraph{Visual Prompt Alignment.}
We perform the \texttt{Align} operation to localize target objects in the input images, by visual prompts. 
To handle flexible visual prompts, which can be image descriptions of objects in any view, scene, style, or timestamp, we employ contrastive mechanisms \cite{wu2018unsupervised,he2020moco} as the key design to retrieve target objects at different styles. 
This allows the module to detect the target objects effectively, even when they exhibit diverse visual styles and appearances.

\begin{figure}
  \centering
    \includegraphics[width=0.5\textwidth]{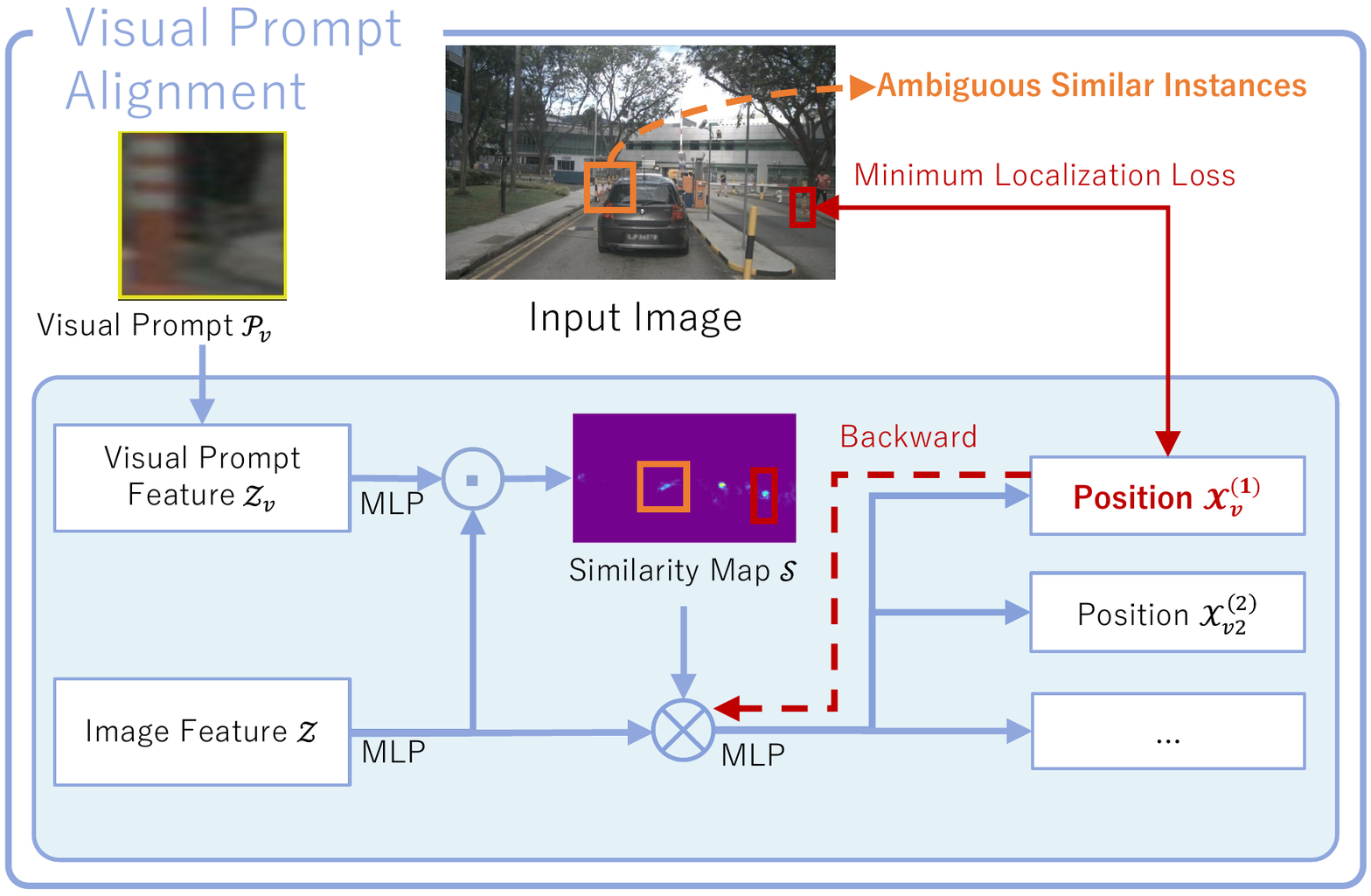}
  \caption{Concrete mechanism of visual prompt alignment. This figure illustrates monocular input. When multi-view images are employed, this alignment operation flattens the different views and still generates $N$ peak candidate positions.}
  \vspace{-0.2in}
  \label{fig:alignhead}
\end{figure}

As described in \Cref{fig:alignhead}, given the visual prompt $\mathcal{P}_v$, we first extract the visual prompt features $\mathcal{Z}_v$ using a lightweight encoder, \eg, ResNet18 \cite{ResNet} in our implementation. Then, we use two separate 2-layer perceptrons, each with 128 and 64 output channels, to align the channel dimensions of the prompt features and the input image features. This results in the aligned feature maps $\hat{\mathcal{Z}_v} \in \mathbb{R}^{M \times 64}$ and $\hat{\mathcal{Z}} \in \mathbb{R}^{KHW \times 64}$, where $M$ is the number of visual prompts, $K$ is the number of image views, $H$ and $W$ are the spatial dimensions of image features. Finally, we compute the cosine similarity ($\odot$ in \Cref{fig:alignhead}) between $\hat{\mathcal{Z}}_v$ and $\hat{\mathcal{Z}}$ to obtain the similarity map $\mathcal{S} \in \mathbb{R}^{M \times KHW}$, which encodes the alignments between the visual prompts and image features.

\myparagraph{Instance Ambiguity \& Loss.}
Sometimes, visual prompts might exhibit instance ambiguity, where multiple objects in the image match the visual descriptions of prompts. 
For example, suppose the visual prompt describes a traffic cone and several similar-looking traffic cones present in the input images (highlighted by ``\textcolor{myorange}{\bf orange box}'' in \Cref{fig:alignhead}).
It can be challenging to uniquely identify the specific object of interest.

For such cases, we design to retrieve all objects with similar identities to the visual prompt. 
Specifically, we modify the align operation to predict multiple spatial coordinates $\mathcal{X}_v = \{ \mathcal{X}^{(i)}_v \}, i \in \{1, 2, ..., N\}$ for each visual prompt, and add the Fourier features of those $N$ coordinates to the visual prompt features to indicate visual prompts in different positions. 
$N$ is set to 4 in our implementation.

For similarity supervision, we generate binary segmentation labels based on the ground-truth 2D boxes. 
Focal loss \cite{FocalLoss} and Dice loss \cite{Milletari2016vnet} are used for optimization. 
To supervise the visual prompt localization $\mathcal{X}_v$, we use the Smooth-$l$1 loss with the target as the center coordinates of ground-truth 2D bounding boxes. 
For dealing with instance ambiguity, we refer to SAM \cite{kirillov2023segany} and only backpropagate the sample with the minimum localization loss during each training iteration.

\myparagraph{Model Design.}
The overall mechanism of \algname-3D Detector is depicted in \Cref{fig:framework} (right).
Based on any traditional offline-trained 3D detector, BEV detector, or monocular detector, we integrate \modelname and train it to be promptable. 
Specifically, \modelname takes features extracted by the image encoder of the corresponding 3D detector, along with various forms of prompts as inputs. 
It encodes these prompts and generates a series of queries, represented as $\mathcal{P} = \{ \mathcal{P}_o; <\mathcal{P}_b, \mathcal{P}_p, \mathcal{P}_v> \}$, where $<...>$ denotes an arbitrary combination of different prompts. 
These queries are then fed into the transformer decoder of the 3D detector to output 3D boxes.

\subsection{Visual Prompt Buffer}
\label{sec:sec3.3}

Visual prompt buffer is a queue that stores visual prompts of missed objects during online inference.
In the default setting, these visual prompts are the 2D regions of risky objects in previous frames.
The \algname system is also flexible enough to allow freely uploading visual prompts, which can be either image content from the current scene or customized images sourced from the Internet based on road descriptions or user-specified requests.
This versatility makes the \algname system applicable to a wide spectrum of scenarios.

\myparagraph{Dequeue.}
To prevent unbounded growth, we adopt a \emph{dequeue} policy that combines score-based eviction and redundancy pruning.
\mrev{During training, we mix in \emph{negative} visual prompts—patches that do not correspond to any target in the current frame (e.g., background crops or off-timestamp instances)—and supervise the model to assign them low confidence.}
\mrev{This calibration causes non-existent or contradictory prompts to be naturally down-weighted at test time, enabling confidence-based removal.}
Concretely, we first filter out prompts whose associated detections consistently fall below a confidence threshold.
We then suppress near-duplicates by measuring intersection-over-union (IoU) between predictions and removing redundant prompts.
This dynamic maintenance balances latency and accuracy by adaptively incorporating informative prompts while pruning stale or duplicate ones online.


\section{Experiments}
We proceed to evaluate our \algname system from two key aspects:
\begin{itemize} [itemsep=0pt, topsep=2pt]
    \item How does \algname perform with various adapted offline 3D detectors for instant online error rectification? 
    \item How does \algname enhance offline-trained 3D detectors when faced with out-of-training-distribution scenarios?
\end{itemize}

\subsection{Experimental Setup}
\label{sec:sec4.1}

\myparagraph{Dataset.}
Experiments are done on the nuScenes dataset \cite{nuscenes2019} with 1,000 autonomous driving sequences, one of the most popular datasets for autonomous driving research. 

\myparagraph{Tasks.} 
Experiments are conducted in two aspects to answer the questions above.
The first experiment aims to test the \algname system in boosting the performance of offline-trained 3D detectors via online error rectification.
We conduct this verification by applying \algname system to various established offline 3D detectors~\cite{zhang2022monodetr,li2022bevformer,yang2023bevformerV2,liu2024ray,wang2023streampetr, sparse4dv2,wang2023object}. 

Then, we validate the \algname system to correct detection errors in out-of-training-distribution scenarios. To conduct a thorough quantitative analysis of this aspect, we established four tasks under different settings: \textbf{(a.)} discarding 80\% labels of distant objects farther than 30m during training, and test the error corrections for detecting distant objects; \textbf{(b.)} discarding 80\% labels of instances labeled as vehicle, including ``car'', ``truck'', ``C.V.'', ``bus'', and ``trailer'', and test the error correction in vehicles; \textbf{(c.)} discarding all annotations of class ``truck'' and ``bus'', and test the zero-shot ability of \algname on those discarded classes; \textbf{(d.)} discarding all training data of the scenario of ``Nighttime'' and ``Rainy'', and test the improvements of \algname on scenarios with domain gap. 
For these tasks, we base \algname system on the monocular algorithm, MonoDETR~\cite{zhang2022monodetr}, and test it under each setting separately, as the monocular setting demonstrates the most general applicability.
Based on \algname-MonoDETR, we also present qualitative results to show the potential of \algname to address corner cases in challenging real-world scenarios through auxiliary feedback during test time.
Through these experiments, we demonstrate \algname as an effective system to adapt offline 3D detectors to challenging scenarios.

\myparagraph{Entity Detection Score.}
As we focus on enabling existing 3D detectors to rectify missing detections, both in-distribution and out-of-distribution, we adopt a class-agnostic setting. We remove all class annotations of 3D objects during training, treating all objects as entities \cite{kirillov2023segany,qi2022openworld}. For evaluation, we use the Entity Detection Score (EDS), a class-agnostic version of the nuScenes Detection Score (NDS).

\myparagraph{Implementation Details.}
We implement our method based on the MMDetection3D codebase \cite{mmdet3d2020}, and conduct all experiments on a server with 8$\times$ A100 GPUs. 
In \modelname, we use a ResNet18 \cite{ResNet} to extract visual prompt features. All visual prompts are resized to $224 \times 224$ before being sent to \modelname. 
For training, we use AdamW \cite{adam,adamw} with a batch size of 16. 
We initialize the learning rate as 2e-4, adjusted by the cosine annealing policy. 
When training point and box prompts, we simulate user inputs with noise by adding perturbations to the ground truth. To ensure the visual prompts are robust across multiple scenes, timestamps, and styles, we choose visual prompts of target objects not only from the image patches of the current frame but also randomly from previous and future frames within a range of $\pm5$. 
Flip operations are used as data augmentations. 
During testing, we mimic the process of auxiliary interventions. 
For user feedback, we simulate online missed feedback by comparing the distance between the ground truth and detected 3D boxes. Any ground truth without a detection result within 2 meters is treated as a missed instance and added to the prompt buffer.
For road description, we use web-derived visual prompts as the input and keep a fixed prompt buffer for error correction. 
For feedback from the 2D detector, we compare the 3D detection results with the Grounding DINO~\cite{liu2024grounding} 2D detection results. Any mismatch with an IoU (Intersection over Union) below 0.5 is considered a misalignment and used as an input for feedback.
\mrev{All experiments are conducted under continuous online operation: the
validation clips were concatenated and inference proceeded sequentially from one clip to the next, simulating persistent
 deployment.}
We remove redundant predictions by non-max-suppression (NMS) with the IoU threshold of 0.5. 
%
To more closely match real-world usage rather than simply optimizing for higher metric scores, we set the classification confidence threshold to 0.3 in all experiments unless otherwise specified.

\begin{table*}[htbp]
\centering
\caption{\mrev{\textbf{Effect of \algname\ on various 3D detectors.} \algname\ improves test-time performance of offline-trained detectors during online inference without extra training. 
To compare directly with the baselines’ official scores, we set the confidence threshold to 0.01.
For false-alarm analysis, we report \emph{FP/frame} and \emph{FN/frame}.
}}
\label{tab:main}
\tablestyle{10pt}{1.4}
\resizebox{\textwidth}{!}{
\begin{tabular}{l|c|c|cc|cc}
\toprule[1.5pt]
Method  & Backbone & Type  
& mAP (\%) $\uparrow$ & EDS (\%) $\uparrow$
& \mrev{FP/frame@0.01 $\downarrow$} & \mrev{FN/frame@0.01 $\downarrow$} \\
\midrule\midrule

MonoDETR \cite{zhang2022monodetr} 
& \multirow{2}{*}{R101} & \multirow{2}{*}{Monocular}
& 42.6 & 50.1 & \mrev{49.9} & \mrev{1.2} \\
\algname-MonoDETR 
&  & 
& \cellcolor{LightCyan} 45.0  (\textcolor{myred}{\textbf{+2.4}})
& \cellcolor{LightCyan} 51.7 (\textcolor{myred}{\textbf{+1.6}})
& \mrev{49.9} & \mrev{0.9} \\
\midrule

MV2D~\cite{wang2023object}      
& \multirow{2}{*}{R50} & \multirow{2}{*}{Multiview}
& 43.0 & 52.8 & \mrev{32.2} &\mrev{2.0}  \\
\algname-MV2D 
&  & 
& \cellcolor{LightCyan} 52.7 (\textcolor{myred}{\textbf{+9.7}})
& \cellcolor{LightCyan} 57.2 (\textcolor{myred}{\textbf{+4.4}})
& \mrev{32.2} & \mrev{0.8} \\
\midrule

Sparse4Dv2~\cite{sparse4dv2}     
& \multirow{2}{*}{R50} & \multirow{2}{*}{Multiview + Temporal}
& 44.4 & 53.2 & \mrev{42.7}  & \mrev{1.9} \\
\algname-Sparse4Dv2 
&  & 
& \cellcolor{LightCyan} 54.2 (\textcolor{myred}{\textbf{+9.8}})
& \cellcolor{LightCyan} 57.8 (\textcolor{myred}{\textbf{+4.6}})
& \mrev{42.7} & \mrev{0.7} \\
\midrule

BEVFormer~\cite{li2022bevformer}      
& \multirow{2}{*}{R101} & \multirow{2}{*}{BEV + Temporal}
& 41.1 & 50.2 & \mrev{34.2} & \mrev{1.9} \\
\algname-BEVFormer 
&  & 
& \cellcolor{LightCyan} 50.2 (\textcolor{myred}{\textbf{+9.1}})
& \cellcolor{LightCyan} 54.4 (\textcolor{myred}{\textbf{+4.2}})
& \mrev{34.4} & \mrev{0.8} \\
\midrule

BEVFormerV2-t8~\cite{yang2023bevformerV2} 
& \multirow{2}{*}{R50} & \multirow{2}{*}{BEV + Temporal}
& 45.6 & 54.1 & \mrev{42.2} & \mrev{1.7} \\
\algname-BEVFormerV2-t8 
&  & 
& \cellcolor{LightCyan} 54.9 (\textcolor{myred}{\textbf{+9.3}})
& \cellcolor{LightCyan} 58.6 (\textcolor{myred}{\textbf{+4.5}})
& \mrev{42.3} &  \mrev{0.6} \\
\midrule

RayDN~\cite{liu2024ray}          
& \multirow{2}{*}{R50} & \multirow{2}{*}{BEV + Temporal}
& 44.9 & 53.6 & \mrev{39.8} & \mrev{1.8} \\
\algname-RayDN 
&  & 
& \cellcolor{LightCyan} 54.4 (\textcolor{myred}{\textbf{+9.5}})
& \cellcolor{LightCyan} 58.1 (\textcolor{myred}{\textbf{+4.5}})
& \mrev{39.8} & \mrev{0.7} \\
\midrule

StreamPETR~\cite{wang2023streampetr}     
& \multirow{2}{*}{V2-99} & \multirow{2}{*}{BEV + Temporal}
& 45.0 & 53.8 & \mrev{42.3} & \mrev{1.8} \\
\algname-StreamPETR 
&  & 
& \cellcolor{LightCyan} 51.6 (\textcolor{myred}{\textbf{+6.6}})
& \cellcolor{LightCyan} 55.6 (\textcolor{myred}{\textbf{+1.8}})
& \mrev{42.4} & \mrev{0.7}  \\
\bottomrule[1.5pt]
\end{tabular}
}
\end{table*}

\begin{table*}[htbp]
    \caption{\textbf{Experiments on out-of-training-distribution scenarios.} TTC system achieves substantial gains with limited or even no labels under different challenging test cases.}
    \label{tab:zero_shots_all}
    \vspace{-0.1cm}
    \tablestyle{3.5pt}{1.1}
    \centering
        \begin{subtable}[h]{0.48\textwidth}
        \centering
        \caption{\textbf{Long-range rectification.} Effect of \algname system in detecting distant objects with 20\% annotations.}
        \label{tab:zero_shots_distant}
        \resizebox{\linewidth}{!}{
            \begin{tabular}{l|cc|cc}
            \toprule[1.5pt]
            \multicolumn{1}{l|}{ \multirow{2}{*}{\makecell[l]{Model \\ Setting }}} & 
            \multicolumn{2}{c|}{ \multirow{1}{*}{All (0m-Inf)}} & 
            \multicolumn{2}{c}{ \multirow{1}{*}{Dist. (30m-Inf)}} \\
            & 
            mAP (\%) & EDS (\%) & 
            mAP (\%) & EDS (\%) \\
            \midrule
            Point & 41.6 & 52.5 & 17.8 & 35.0 \\
            Box & 44.3 & 54.3 & 19.2  & 36.3 \\
            Visual & 42.6  & 53.0  & 18.3 &  36.1\\
            \midrule
            \rowcolor{LightCyan} MonoDETR & 31.4 & 47.6 & 0.0 & 0.0 \\
            \rowcolor{LightCyan} \algname-MonoDETR & \textbf{39.6} & \textbf{52.1} & \bf 10.1 & \bf 27.9 \\
            \rowcolor{LightCyan} \multicolumn{1}{c|}{ \multirow{1}{*}{\textcolor{myred} {\textbf{$\Delta$}}}} & \textcolor{myred}{\textbf{+8.2}} & \textcolor{myred}{\textbf{+4.5}} & \textcolor{myred}{\textbf{+10.1}} & \textcolor{myred}{\textbf{+27.9}}\\
            \bottomrule[1.5pt]
            \end{tabular}
        }
        \end{subtable}
    \hfill
        \begin{subtable}[h]{0.48\textwidth}
        \centering
        \caption{\textbf{Vehicle-focused rectification.} Effect of  \algname system on vehicle objects with 20\% annotations.}
        \label{tab:zero_shots_vehicle}
        \resizebox{\linewidth}{!}{
            \begin{tabular}{l|cc|cc}
            \toprule[1.5pt]
            \multicolumn{1}{l|}{ \multirow{2}{*}{\makecell[l]{Model \\ Setting }}} & 
            \multicolumn{2}{c|}{ \multirow{1}{*}{All}} & 
            \multicolumn{2}{c}{ \multirow{1}{*}{Vehicle}} \\
            & 
            mAP (\%) & EDS (\%) & 
            mAP (\%) & EDS (\%) \\
            \midrule
            Point & 38.0 & 48.2 & 33.5 & 52.8 \\
            Box & 41.2 & 50.3 & 36.7 & 54.4 \\
            Visual & 39.2 & 47.7 & 34.6 & 51.8 \\
            \midrule
            \rowcolor{LightCyan} MonoDETR & 17.6 & 29.0 & 2.4 & 36.0 \\
            \rowcolor{LightCyan} \algname-MonoDETR & \textbf{28.5} & \textbf{35.4} & \textbf{22.4} & \textbf{44.2} \\\rowcolor{LightCyan} \multicolumn{1}{c|}{ \multirow{1}{*}{\textcolor{myred} {\textbf{$\Delta$}}}} & \textcolor{myred}{\textbf{+10.9}} & \textcolor{myred}{\textbf{+6.4}} & \textcolor{myred}{\textbf{+20.0}} & \textcolor{myred}{\textbf{+8.2}} \\
            \bottomrule[1.5pt]
            \end{tabular}
        }
        \end{subtable}
    \\
    \vspace{.1in}
        \begin{subtable}[h]{0.48\textwidth}
        \centering
        \caption{\textbf{Novel object rectification.} Effect of \algname system on objects of novel classes unseen in the training set.}
        \label{tab:zero_shots_novelclass}
        \resizebox{\linewidth}{!}{
            \begin{tabular}{l|cc|cc}
            \toprule[1.5pt]
            \multicolumn{1}{l|}{ \multirow{2}{*}{\makecell[l]{Model \\ Setting }}} & 
            \multicolumn{2}{c|}{ \multirow{1}{*}{All}} & 
            \multicolumn{2}{c}{ \multirow{1}{*}{Unseen}} \\
            & 
            mAP (\%) & EDS (\%) & 
            mAP (\%) & EDS (\%) \\
            \midrule
            Point & 35.2 & 48.7 & 11.9 &33.3  \\
            Box & 38.5 & 50.9 & 17.0 & 36.6 \\
            Visual & 35.4 & 47.3 & 11.2 & 33.3 \\
            \midrule
            \rowcolor{LightCyan} MonoDETR & 28.3 & 47.0 & 0.0 & 0.0 \\
            \rowcolor{LightCyan} \algname-MonoDETR & \bf 34.6 & \bf 49.0 & \bf 7.2 & \bf 30.8 \\\rowcolor{LightCyan} \multicolumn{1}{c|}{ \multirow{1}{*}{\textcolor{myred} {\textbf{$\Delta$}}}} & \textcolor{myred}{\textbf{+6.3}} & \textcolor{myred}{\textbf{+2.0}} & \textcolor{myred}{\textbf{+7.2}} & \textcolor{myred}{\textbf{+30.8}}\\
            \bottomrule[1.5pt]
            \end{tabular}
        }
        \end{subtable}
    \hfill
        \begin{subtable}[h]{0.48\textwidth}
        \centering
        \caption{\textbf{Domain shift rectification.} Effect of \algname system on objects in scenarios with domain gap.}
        \label{tab:zero_shots_domaingap}
        \resizebox{\linewidth}{!}{
            \begin{tabular}{l|cc|cc}
            \toprule[1.5pt]
            \multicolumn{1}{l|}{ \multirow{2}{*}{\makecell[l]{Model \\ Setting }}} & 
            \multicolumn{2}{c|}{ \multirow{1}{*}{All}} & 
            \multicolumn{2}{c}{ \multirow{1}{*}{Rain \& Night}} \\
            & 
            mAP (\%) & EDS (\%) & 
            mAP (\%) & EDS (\%) \\
            \midrule
            Point & 39.0 & 49.4  & 30.5 & 44.3 \\
            Box & 42.6 & 51.8  & 33.4 & 46.3 \\
            Visual & 39.8 & 48.6 & 29.4 & 43.0 \\
            \midrule
            \rowcolor{LightCyan} MonoDETR & 34.5 &  48.4 & 25.2 & 43.1 \\
            \rowcolor{LightCyan} \algname-MonoDETR & \bf 39.5 & \bf 51.8 & \textbf{29.0} & \textbf{44.5} \\\rowcolor{LightCyan} \multicolumn{1}{c|}{ \multirow{1}{*}{\textcolor{myred} {\textbf{$\Delta$}}}} & \textcolor{myred} {\textbf{+5.0}} & \textcolor{myred} {\textbf{+3.4}}  & \textcolor{myred} {\textbf{+3.8}} & \textcolor{myred} {\textbf{+1.4}} \\
            \bottomrule[1.5pt]
            \end{tabular}
        }
        \end{subtable}
    \vspace{-.1in}
\end{table*}
\begin{figure*}[t]
	\centering
        \includegraphics[width=0.92\linewidth]{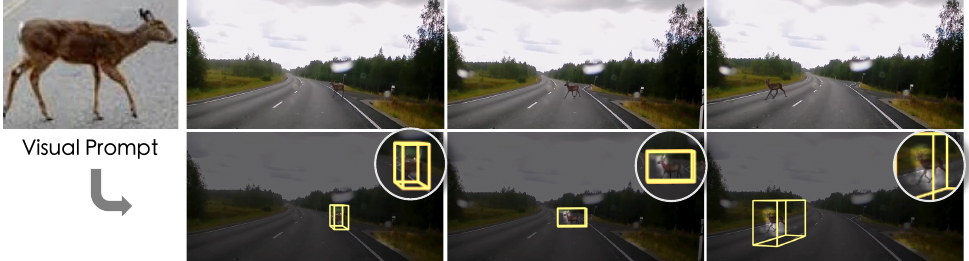}
 \caption{\textbf{Qualitative visualization of real-world scenes (collected from YouTube).} We visualize the zero-shot 3D detection results in a real-world scenario. In this case, the prompt buffer contains a visual prompt of a deer. Higher responses from the visual prompt alignment are highlighted by brighter colors. As shown, although \textbf{trained solely on nuScenes}, \algname system can still accurately localize ``unseen'' objects in the input image. Best viewed in color.}
	\label{fig:viz_prompt_geer}
\end{figure*}

\subsection{Main Result}

\myparagraph{Effectiveness of \algname in Test-time Error Correction.}
Test-time error correction ability without re-training is the core capability of the \algname system. 
We verify this by incorporating traditional offline-trained 3D detectors into the \algname system and compare the performance without test-time parameter updates.
For thorough verification, we select various offline-trained 3D detectors, including monocular \cite{zhang2022monodetr}, multi-view \cite{wang2023object,sparse4dv2}, and BEV ones \cite{li2022bevformer,yang2023bevformerV2,wang2023streampetr,liu2024ray}. 
The online feedback is derived from the human clicks
which is then converted into visual prompts. The feedback interval is set to 5 frames.
As shown in \Cref{tab:main}, the \algname system substantially improves the test-time performance of offline-trained 3D detectors, \eg, 9.1\% and 6.6\% mAP improvements on BEVFormer and StreamPETR, without requiring additional training after deployment. These demonstrate the effectiveness of the \algname system in instantly correcting test-time errors during online inference.

\myparagraph{Effectiveness of \algname in Out-of-Training Scenarios.}
\Cref{tab:zero_shots_all} presents results to validate the \algname system in challenging and extreme scenarios. 
In these experiments, we base \algname system on MonoDETR for its simple monocular setting. ``Point'', ``Box'', and ``Visual'' in \Cref{tab:zero_shots_all} are \algname-MonoDETR directly using point, box, or image patch of target objects in the input image as input prompts, which are impractical unless automated annotation scenario. 
These are performance upper bounds as they receive human feedback in every frame.
Compared with the baseline MonoDETR, ``\algname-MonoDETR'' in the table obtains visual prompts of missing objects from user at a feedback interval of 5 frames, enabling continuous correction without any test-time weight updates.

\Cref{tab:zero_shots_distant} evaluates the effectiveness of \algname system in rectifying detection errors on distant objects (beyond 30m). During training, 80\% of the far-away annotations are removed, resulting in a 3D detector with poor long-range performance (0.0\% mAP and EDS). Powered by the \algname system, the performance on these distant objects is instantly improved to 10.1\% mAP and 27.9\% EDS, without any extra training.
The experiment is then extended to all vehicles in the nuScenes dataset, as shown in \Cref{tab:zero_shots_vehicle}. The \algname achieves impressive performance gains of 20.0\% mAP and 8.2\% EDS.

Furthermore, in \Cref{tab:zero_shots_novelclass} and \Cref{tab:zero_shots_domaingap}, we evaluate the \algname system in two challenging scenarios: encountering unseen objects not present in the training data; handling domain shifts, such as transitioning from sunny to rainy or nighttime conditions. In \Cref{tab:zero_shots_novelclass}, the \algname system shows effectiveness in successfully detecting novel objects not labeled in the training set, achieving 7.2\% mAP and 30.8\% EDS on these novel objects, significantly improving the offline-trained baseline with 0.0\% mAP and EDS. In \Cref{tab:zero_shots_domaingap}, we find \algname also works well for online error rectification when driving into scenarios with domain shifts, providing 3.8\% mAP and 1.4\% EDS improvements.

Regarding qualitative results, \Cref{fig:viz_prompt_geer} further illustrates a case of the zero-shot capability in the real world. Despite being trained solely on the nuScenes dataset, \algname-MonoDETR can detect a \texttt{Deer} using a visual prompt of another deer from a different viewpoint (simulating an internet image search for road description “deers might appear on this road”), which is challenging for traditional offline detectors. More qualitative examples can be found in Appendix~\Cref{sec:supp_vis}.

These experiments demonstrate \algname as an effective and versatile system for instant error correction, excelling at handling missing distant objects, unseen object categories, and domain shifts, without requiring any additional training.
The superior performance of \algname system in these challenging real-world scenarios highlights its potential to enable robust and adaptable 3D object detection systems.

\subsection{Robustness of Visual Prompts}
\label{sec:4.3}

As a crucial component of the continuous test-time error correction system, we conduct a series of experiments on visual prompts in this section.
Specifically, we analyze the robustness of visual prompts against: \textit{(i)} variations in prompt sources and styles, \textit{(ii)} changes in viewpoints across frames, \textit{(iii)} spatial perturbations in provided prompts, \textit{(iv)} reduced feedback frequency, \mrev{\textit{(v)} noisy (off-topic) visual prompts, and \textit{(vi)} rapidly changing prompts.}

\begin{figure*}[htbp]
    \centering
     \begin{minipage}{0.48\textwidth}
        \centering
      \includegraphics[width=\linewidth]{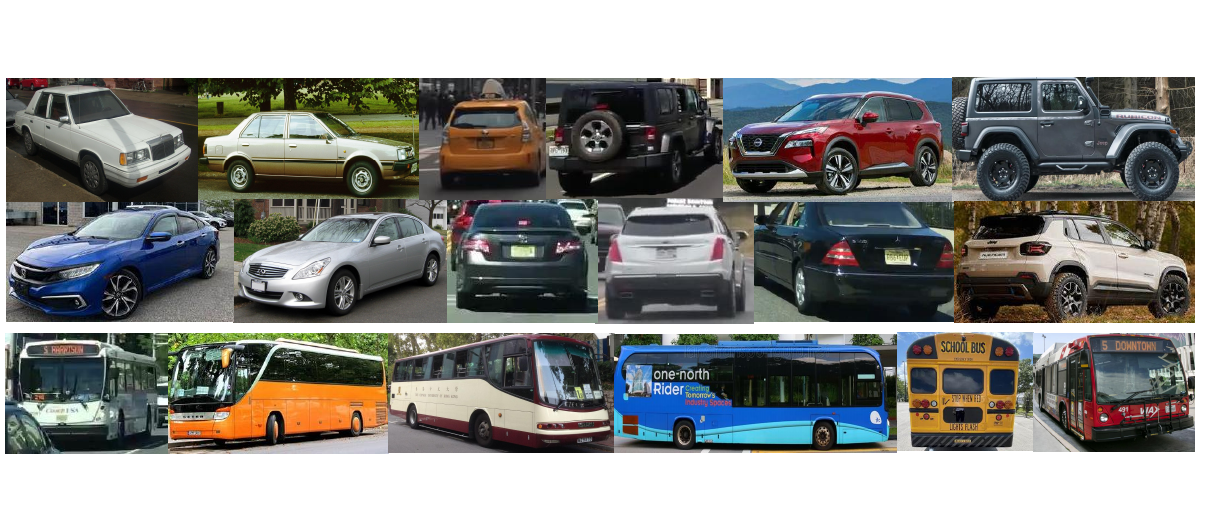} 
        \caption{\textbf{Visual prompt examples derived from the Internet.} We select 12 cars and 6 buses with different views and colors from websites to cover the distribution of object appearances in the nuScenes dataset comprehensively.}
        \label{fig:web_prompts_sample}
    \end{minipage}\hfill
    \begin{minipage}{0.48\textwidth}
        \centering
        \captionof{table}{\textbf{Results with web-derived visual prompts.} \algname-MonoDETR$^\dag$ indicates the model with frozen prompt buffer containing pre-assigned visual prompts derived from the Internet. \algname system can still significantly improve traditional 3D detectors even using web prompts in scenarios with very limited labeled data.}
        \label{tab:web_prompts}
        \resizebox{\textwidth}{!}{
        \begin{tabular}{l|c|cc}
        \toprule
        Category & Method  & mAP (\%) $\uparrow$ & EDS (\%) $\uparrow$ \\ 
        \midrule
        \midrule
        \multirow{2}{*}{Car}    & MonoDETR & 3.2 &  37.5 \\
        & \cellcolor{LightCyan}  \algname-MonoDETR$^\dag$ &\cellcolor{LightCyan}  \bf 20.9  (\textcolor{myred}{\textbf{+17.7}}) & \cellcolor{LightCyan}    45.3 (\textcolor{myred}{\textbf{+7.8}}) \\
        \midrule
        \multirow{2}{*}{Bus}    & MonoDETR & 0.4 & 28.1  \\
        &  \cellcolor{LightCyan}  \algname-MonoDETR$^\dag$ & \cellcolor{LightCyan} \bf 17.4  (\textcolor{myred}{\textbf{+17.0}}) & \cellcolor{LightCyan} \bf   40.2 (\textcolor{myred}{\textbf{+12.1}}) \\
        \bottomrule
    \end{tabular}
    }
    \end{minipage}
      
\end{figure*}
\begin{figure*}[htbp]
	\centering
        \includegraphics[width=0.96\linewidth]{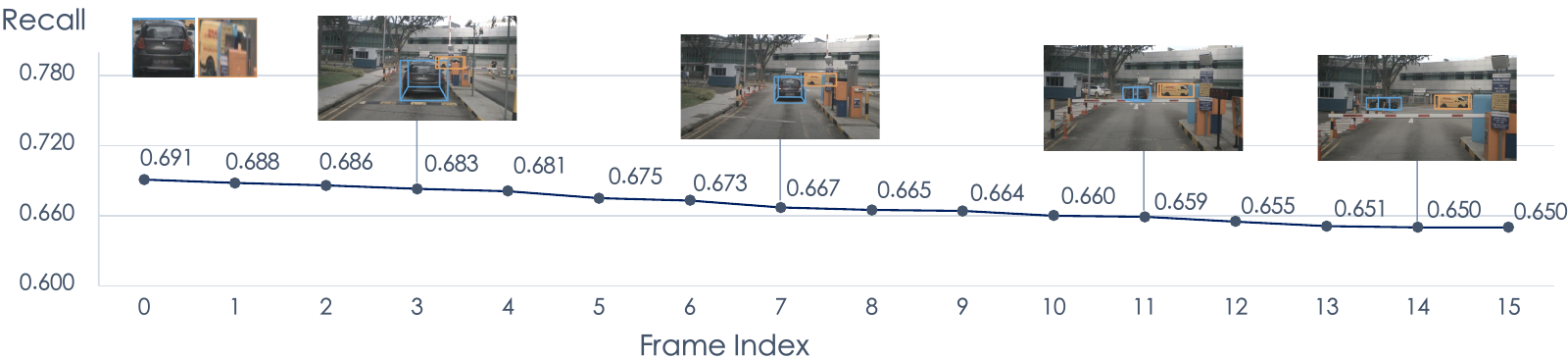}
	\caption{\textbf{Experiments on visual prompts from arbitrary temporal frames.} \algname-3D Detectors can effectively locate and detect target objects in future frames, using its image patch at Frame \#0. 
 }
	\label{fig:exp_visual_prompt}
	\vspace{-0.2in}
\end{figure*}

\myparagraph{Robustness over Web-derived Visual Prompts.}
Visual prompts can be arbitrary imagery views of target objects and can be from any image source and styles. For example, we can use images sourced from the Internet as visual prompts. We fix the prompt buffer with visual prompts from the Internet during inference, and assess the effectiveness of handling visual prompts with diverse styles.
We employ the model from \Cref{tab:zero_shots_vehicle} (\algname-MonoDETR trained with 20\% labels of vehicles), and select 12 car and 6 bus images from the Internet, which resemble those in the nuScenes dataset, as visual prompts for online correction. We show them in \Cref{fig:web_prompts_sample}.
As listed in \Cref{tab:web_prompts}, \algname system demonstrates strong online correction capabilities. Even with prompts sourced from the Internet with various styles and poses, \algname system still improves the offline-trained baseline by 17.7\% mAP on ``Car'' objects under this extremely challenging setting. This further underscores the robustness of \algname system and highlights its potential to address the long-tail challenges in real-world scenarios.

\myparagraph{Robustness over Arbitrary Visual Prompt Views.}
We also investigate the capability of \algname system in handling visual prompts of target objects across scenes and times.
In \Cref{fig:exp_visual_prompt}, we study whether our system can successfully associate objects with their visual prompts from arbitrary frames.
Specifically, for each video clip of the nuScenes validation set, we use the image patch of target objects in the first frame as visual prompts, then detect and track target objects in subsequent frames, and compute the recall rate for evaluation with arbitrary views. \Cref{fig:exp_visual_prompt} presents the results of \algname-MonoDETR, showing that despite significant differences in viewpoints and object poses between frames, the recall rate does not drop dramatically as the ego vehicle moves.
This highlights the robustness of the \algname detectors in handling visual prompts with arbitrary views across scenes and times, indicating that the \algname system can effectively process feedback that may involve temporal delays.

\begin{figure}
  \centering
  \vspace{-0.3in}
  \includegraphics[width=0.486\textwidth]{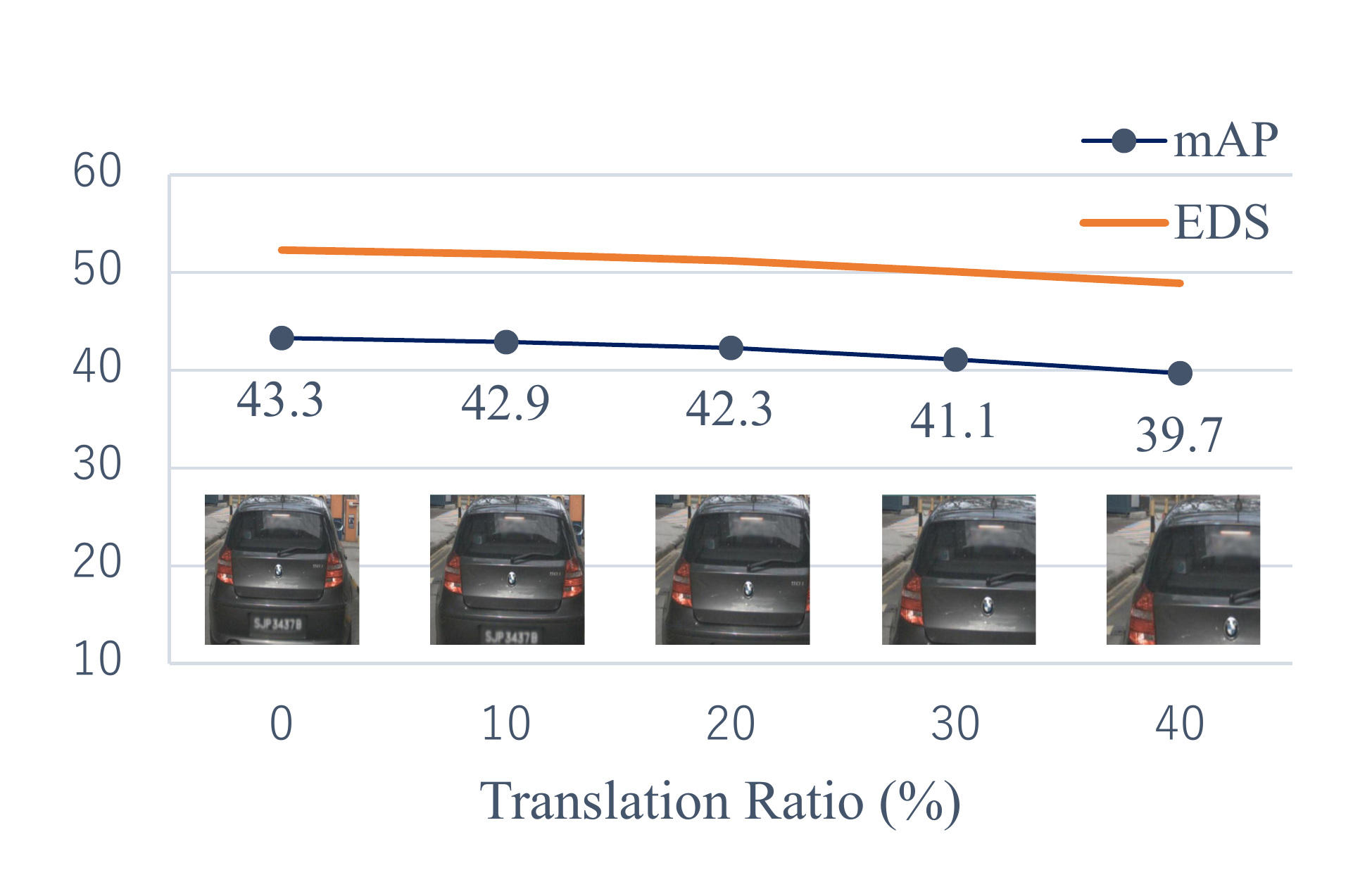}
  \vspace{-0.15in}
  \caption{\textbf{Robustness over positional perturbations.} We evaluate \algname-MonoDETR under controlled perturbations during test-time inference. Specifically, we introduce translation offsets to the ground-truth 2D centers with maximum ratios of 0\%, 10\%, 20\%, 30\%, and 40\%.
}
  \vspace{-0.12in}
  \label{fig:pos_offset_prompts}
\end{figure}

\myparagraph{Robustness over Positional Perturbations.}
Since visual prompts during online inference are derived from user clicks or 2D detectors, they may deviate from the precise 2D center of the target object. 
To analyze the impact of such positional shifts, we introduce controlled perturbations during test-time inference.
Specifically, we apply translation offsets to the ground-truth 2D centers, varying the maximum displacement ratio from 0\% to 40\%, and evaluate the system’s robustness under these conditions.
As shown in \Cref{fig:pos_offset_prompts}, despite increasing perturbations, \algname-MonoDETR exhibits only minor performance degradation, demonstrating strong resilience to positional variations. This underscores the robustness of the \algname system in handling various interactions, where precise locations may not always be guaranteed.

\myparagraph{Robustness over Visual Prompt Collection Frequency.}
In practical deployment, human feedback and open-vocabulary 2D detectors (\eg, Grounding DINO) are often delayed, making it infeasible to provide corrections for every missed detection in real time. To evaluate the robustness of \algname under varying feedback frequencies, we design experiments that simulate different feedback intervals.
\begin{table}
    \centering
    \caption{\textbf{mAP comparisons under different feedback collection intervals.} ``N=0'' means feedback is collected every frame, while ``N=2'' means every 2 frames.}
    \label{tab:human_feedback_frequency}
    \small
    \setlength{\tabcolsep}{2mm}  
\begin{tabular}{l|cccc}
\toprule
    Feedback Interval $N$ & 0  & 2 & 4  & 6  \\ 
\midrule
    \algname-BEVFormer  & 47.8 & 47.5 & 47.3 & 46.9 \\
    \algname-Sparse4Dv2  & 53.5 & 53.1 & 52.4 & 51.8 \\ 
    \algname-MV2D  & 51.0 & 50.7  & 50.2  & 49.6 \\ 
\bottomrule
\end{tabular}
\end{table}
As shown in \Cref{tab:human_feedback_frequency}, we update the prompt buffer at different frame intervals to mimic varying levels of feedback availability.
Despite reducing the update frequency (from every frame to every 6 frames), \algname maintains strong error correction performance, demonstrating its resilience to infrequent feedback. This suggests that \algname can effectively operate in real-world scenarios where continuous supervision is impractical.

\myparagraph{\mrev{Robustness over Noisy Visual Prompts.}}
\mrev{To evaluate robustness against completely misleading prompts, we augment the fixed prompt buffer used in \Cref{tab:web_prompts} with two \emph{animal} exemplars (one \textbf{dog}, one \textbf{kangaroo}; prompt images in Appendix~\Cref{fig:app-c5-2}) in addition to vehicle prompts. These classes are virtually absent in nuScenes imagery and thus serve as \emph{off-topic} (noisy) prompts. We evaluate the \textbf{Car} category using \algname-MonoDETR$^\dag$ from \Cref{tab:web_prompts} (trained with 20\% vehicle labels), comparing runs \emph{with} and \emph{without} the added animal prompts. As shown in \Cref{tab:noisy_prompts_animals}, adding these off-topic prompts do not degrade accuracy, further demonstrating TTC’s robustness to noisy prompts.}

\begin{table}[t]
\centering
\small
\setlength{\tabcolsep}{4.6pt}
\renewcommand{\arraystretch}{0.98}
\caption{\mrev{\textbf{Robustness to noisy visual prompts.} Comparison of fixed-buffer \algname-MonoDETR$^\dag$ \emph{with} and \emph{without} off-topic animal prompts (dog \& kangaroo). Evaluation follow the protocol of \Cref{tab:web_prompts}.}}

\label{tab:noisy_prompts_animals}
\begin{tabular}{lcccc}
\toprule
Prompt Type & mAP $\uparrow$ & EDS $\uparrow$ \\
\midrule
Vehicle & 20.9 & 45.3  \\
Vehicle + Animal  & 20.9 & 45.3 \\
\bottomrule
\end{tabular}
\vspace{-2mm}
\end{table}

\myparagraph{\mrev{Robustness over Rapidly Changing Prompts.}}
\mrev{We test \algname under fast ego–motion and frequent occlusions, where prompts appear/disappear quickly. To emulate this regime, we uniformly sub–sample each video by \textbf{50\%}, effectively doubling apparent scene speed and shortening object dwell time, while keeping the 5-frame feedback interval and all thresholds/buffer rules identical to \Cref{tab:main}. As summarized in \Cref{tab:rapid_prompts}, the gains over the base detectors remain \emph{consistently positive} (mAP/EDS and FN/frame), albeit smaller than in the full–frame setting. This indicates that \algname is robust to rapidly changing prompts.}

\begin{table}[t]
\centering
\small
\setlength{\tabcolsep}{4.6pt}
\renewcommand{\arraystretch}{0.98}
\caption{\mrev{\textbf{Robustness to rapidly changing prompts.} We choose BEVFormer as the base detector.}}

\label{tab:rapid_prompts}
\resizebox{0.47\textwidth}{!}{
\begin{tabular}{lcccc}
\toprule
Method & mAP $\uparrow$ & EDS $\uparrow$ & FP/frame $\downarrow$ & FN/frame $\downarrow$ \\
\midrule
BEVFormer & 40.2 & 49.6 & 34.9 & 1.9 \\
\cellcolor{LightCyan}TTC-BEVFormer  & \cellcolor{LightCyan}45.9 (\textcolor{myred}{+5.7}) & \cellcolor{LightCyan}51.9 (\textcolor{myred}{+2.3}) & \cellcolor{LightCyan}35.0  & \cellcolor{LightCyan}1.2 \\
\bottomrule
\end{tabular}
}
\end{table}

\subsection{Additional Comparisons and System Analysis}
\label{sec:4.4}

Beyond evaluating the effectiveness of \algname and the robustness of visual prompts, we conduct additional comparisons to clarify its advantages over alternative approaches: \textit{(i)} comparison with an SOT-based online error–correction baseline and \textit{(ii)} contrast with offline fine–tuning driven by test–time 2D feedback. We also provide system–level analyses: \textit{(iii)} efficiency in terms of model size and inference latency; \mrev{\textit{(iv)} cross–dataset transfer ability; \textit{(v)} feasibility of fully automated feedback based on 2D–3D misalignment; and \textit{(vi)} FP trade-off induced by visual prompts.}

\myparagraph{Comparison with SOT-based Online Error Correction.}
An intuitive way to integrate online error correction is to combine a 2D tracking algorithm with a 3D detection head. To explore this possibility, we implement a baseline using ATOM~\cite{danelljan2019atom} for 2D tracking and MV2D~\cite{wang2023object} as the 3D object detector. MV2D leverages ATOM's tracking results of missing objects from previous frames to initialize queries, mimicking an error correction mechanism.
In this setup, we perform human intervention every 5 frames to provide a list of missing objects along with their 2D bounding boxes for ATOM as the reference.
For a fair comparison, the visual prompts of \algname-MV2D are derived from user-drawn bounding boxes every 5 frames.
As shown in \Cref{tab:oa_vs_sot}, \algname, equipped with the OA module, significantly outperforms this SOT-based approach in error correction. This highlights the limitations of simply combining SOT with a 3D detector and underscores the necessity of a dedicated online correction framework.
\begin{table}
    \centering
    \caption{\textbf{Comparison of error correction performance with SOT + 3D detector head.} The results demonstrate the superior performance of \algname in online error correction, highlighting the limitations of the baseline approach where simply combining SOT with a 3D detection head does not effectively address error correction.}
    \label{tab:oa_vs_sot}
    \small
    \setlength{\tabcolsep}{3.mm}
    \begin{tabular}{l|c|c}
    \toprule
       Method & mAP (\%) $\uparrow$ & EDS (\%) $\uparrow$ \\
        \midrule
        MV2D & 42.6 & 51.9 \\
        \midrule
        SOT + MV2D & 42.1 & 51.2 \\
      \cellcolor{LightCyan}\algname-MV2D & \cellcolor{LightCyan}\bf 50.0 &   \cellcolor{LightCyan}\bf 57.3 \\
    \bottomrule
    \end{tabular}
    \vspace{-2mm}
\end{table}

\myparagraph{Comparison with Offline Fine-tuning on Test-time Feedback.}
An alternative way to leverage auxiliary feedback is offline fine-tuning, where missing objects are collected during inference and later used to fine-tune the model with test-time 2D ground truth. While this enables adaptation, it introduces inherent delays compared to real-time correction.
We evaluate this strategy by fine-tuning MV2D~\cite{wang2023object}, which relies on 2D detection for 3D object inference and can incorporate 2D feedback. In this experiment, we collect human 2D feedback at different intervals and compare MV2D’s fine-tuned performance with \algname-MV2D, where prompts are updated at the same intervals.
Results in \Cref{tab:offline_finetune} show that \algname-MV2D consistently outperforms the fine-tuned MV2D with 2D weak feedback. 
This demonstrates its effectiveness of \algname in real-time streaming applications with only weak feedback feasible.

\begin{table}
    \centering
    \caption{\textbf{mAP comparison between \algname-MV2D and MV2D fine-tuned with 2D feedback annotations.} ``N=0'' means collecting feedback at every frame, while ``N=2'' means every 2 frames. \algname-MV2D consistently outperforms the offline fine-tuned MV2D.}
    \label{tab:offline_finetune}
    \small
    \setlength{\tabcolsep}{2mm}  
\begin{tabular}{l|cccc}
\toprule
    Feedback Interval $N$  & 0  & 2  & 4  & 6 \\ 
\midrule
    MV2D + Offline fine-tune & 43.4 & 42.7  & 41.0 & 39.9 \\  
    \cellcolor{LightCyan}\algname-MV2D  & \cellcolor{LightCyan}\bf 51.0 & \cellcolor{LightCyan}\bf 50.7  & \cellcolor{LightCyan}\bf 50.2  & \cellcolor{LightCyan}\bf 49.6 \\ 
\bottomrule
\end{tabular}
\end{table}

\myparagraph{Efficiency Analysis: Model Size and Latency.}
For online deployment, a system must balance accuracy and efficiency. We analyze \algname’s model size and inference latency, particularly in comparison with recent LLM-based promptable 3D detection methods~\cite{chen2024ll3da,huang2023embodied}. 
\begin{table}
    \centering
    \caption{\textbf{Parameters and latency comparisons} between LLM based 3D detectors, traditional 3D detectors, and related \algname 3D detectors. }
    \label{tab:latency}
    \small
    \setlength{\tabcolsep}{3.5mm}
    \resizebox{0.48\textwidth}{!}{
    \begin{tabular}{l|ccc}
\toprule
    Method  & LL3DA   & MonoDETR & \algname-MonoDETR \\ \midrule
    \#Params $\downarrow$& 118M & 68M & 80M \\
    FPS (Hz) $\uparrow$ & 0.42  & 11.1 & 9.1 \\
\bottomrule
\end{tabular}
}
    \vspace{-0.05in}
\end{table}
Latency in promptable systems arises from two sources: \textit{(i)} response time in providing prompts and \textit{(ii)} the system’s inference speed. While the first factor is unavoidable, the second is crucial for real-time applications.
As shown in \Cref{tab:latency}, LLM-based methods like LL3DA~\cite{chen2024ll3da} exhibit significantly higher latency, making them impractical for autonomous driving. In contrast, \algname introduces minimal overhead compared to its base detector while maintaining real-time performance, confirming its suitability for online deployment in safety-critical scenarios.

\myparagraph{\mrev{Cross-dataset Transfer Analysis.}}
\mrev{We complement the results with a cross-dataset study from nuScenes to Waymo. 
Using MonoDETR as the base detector, \Cref{tab:cross_dataset} shows that direct transfer causes a severe collapse (42.6 $\to$ 0.7 mAP). Accordingly, directly applying TTC brings only a marginal gain of +0.3 mAP.
As discussed in \textit{Cross-Dataset Sensor Alignment}~\cite{zheng2023cross}, for vision-only 3D detectors the dominant factor is the mismatch in camera intrinsics/extrinsics: when moving from nuScenes to Waymo, objects at comparable 3D distance/shape occupy markedly different 2D pixel scales due to focal-length disparities and rig configurations (Appendix~\Cref{fig:cross_dataset}), which induces depth bias.}

\mrev{To decouple this confound, the intrinsic synchronization (\textbf{K-sync}) strategy~\cite{zheng2023cross} is adopted: images are resized and cropped to normalize focal lengths prior to inference. 
Under K-sync, TTC yields a substantial improvement over the K-sync–only baseline on Waymo (\,+3.3 mAP\,), indicating that once intrinsics are aligned, the proposed prediction-level rectification generalizes across datasets without updating model weights (\Cref{tab:cross_dataset}).}

\begin{table}[t]
\centering
\small
\setlength{\tabcolsep}{8pt}
\caption{\mrev{\textbf{mAP results of cross-dataset validation.} The model is trained on nuScenes and evaluated on the nuScenes and Waymo. MonoDETR is used as the base detector.}}

\label{tab:cross_dataset}
\begin{tabular}{lcc}
\toprule
\textbf{Method} & \textbf{nuScenes} & \textbf{Waymo} \\
\midrule
MonoDETR & 42.6 & 0.7 \\
\cellcolor{LightCyan}TTC-MonoDETR & \cellcolor{LightCyan} 45.0 & \cellcolor{LightCyan} 1.0 \\
\midrule
K-sync + MonoDETR & 42.8 & 30.8 \\
\cellcolor{LightCyan}K-sync + TTC-MonoDETR & \cellcolor{LightCyan} \textbf{45.8 (\textcolor{myred}{+3.0})} & \cellcolor{LightCyan} \textbf{34.1 (\textcolor{myred}{+3.3})} \\
\bottomrule
\end{tabular}
\end{table}

\myparagraph{\mrev{Fully Automated Feedback Analysis.}}
\mrev{To address the practical difficulty of obtaining clicks/boxes while driving, we evaluate a \emph{most realistic} deployment mode in which all prompts are generated automatically (no human inputs). 
We harvest feedback from the discrepancy between a 2D detector and the image reprojection of the current 3D predictions. 
Concretely, we consider two 2D detectors: (i) off-the-shelf Grounding DINO~\cite{liu2024grounding}, and (ii) a YOLOv8~\cite{ultralytics_yolov8} model retrained on the nuScenes 2D taxonomy. 
For each frame, every 2D box is compared against the projected 3D boxes; unmatched cases (IoU $<0.5$; optionally gated by reprojection residual) are converted into instance-specific visual prompts by cropping the corresponding image patch and enqueuing it into the buffer. 
Under the same in-domain protocol as \Cref{tab:main}, automated prompts yield consistent improvements in mAP/EDS (\Cref{tab:auto_feedback}).}

\begin{table}[t]
\centering
\small
\setlength{\tabcolsep}{6pt}
\renewcommand{\arraystretch}{0.98}
\caption{\mrev{\textbf{Automated feedback results.} 
2D–3D misalignment is computed between 2D boxes (Grounding DINO / YOLOv8-nuScenes) and the image reprojection of 3D predictions; unmatched 2D boxes are converted into visual prompts.}}
\label{tab:auto_feedback}
\begin{tabular}{lcc}
\toprule
Method & mAP $\uparrow$ & EDS $\uparrow$ \\
\midrule
BEVFormer                       & 41.1   & 50.2   \\
\cellcolor{LightCyan}TTC-BEVFormer (GDINO)   & \cellcolor{LightCyan}41.8 (\textbf{\textcolor{myred}{+0.7}})  & \cellcolor{LightCyan}50.4 (\textbf{\textcolor{myred}{+0.2}})  \\
\cellcolor{LightCyan}TTC-BEVFormer (YOLOv8)  & \cellcolor{LightCyan}42.4 (\textbf{\textcolor{myred}{+1.3}})  & \cellcolor{LightCyan}50.8 (\textbf{\textcolor{myred}{+0.6}})   \\
\bottomrule
\end{tabular}
\vspace{-2mm}
\end{table}

\myparagraph{\mrev{FP Impact Analysis.}}
\mrev{Because visual prompts introduce additional hypotheses at inference, a natural concern is whether they inflate false alarms. We therefore quantify the effect of \algname on false positives in~\Cref{tab:main}. Concretely, we report \emph{FP/frame} and \emph{FN/frame} over the full validation stream, where
$\text{FP/frame}=\tfrac{\text{FP}_{\text{total}}}{\text{frames}}$ and $\text{FN/frame}=\tfrac{\text{FN}_{\text{total}}}{\text{frames}}$.
A detection is matched using the same center–distance protocol as mAP calculation (averaged over $\{0.5,1.0,2.0,4.0\}$\,m).
Across diverse backbones, \algname\ markedly reduces \emph{FN/frame}, while \emph{FP/frame} changes only marginally (\textbf{$\leq\!+0.2$}), indicating that the recall gains do not come at the cost of inflating false alarms. 
We also evaluate a deployment-oriented setting with a confidence threshold of $0.3$; the \emph{FP/frame} is not statistically significant before vs.\ after applying TTC (see Appendix~\Cref{tab:fp_fn_tau03}).}

\mrev{\textit{Why FP remains controlled.} 
As stated in our method, each prompt spawns up to $N$ hypotheses ($N{=}4$). For clarity,
we define a valid prompt is one whose associated object is present in the current scene, whereas an invalid prompt is
stale or contradictory (the object has left the scene or has no corresponding instance). 
On TTC–BEVFormer, we observe on average only $3.9$ visual prompts per frame; each valid prompt yields 1.6 above-threshold candidates on average, and each invalid prompt yields 0.3, indicating that visual prompts do not flood the detector and that invalid prompts are usually low-confidence.
For above-threshold candidates, the outcomes are: 
(i) correct match to the intended object (desired);
(ii) match to a visually similar object that has already been detected—suppressed by NMS, so no extra box appears;
(iii) match to a similar object that is not detected—this becomes a TP for that object rather than an FP;
(iv) fall on an empty region but overlap an FP from the dense query results—removed by NMS, thus not affecting the final FP count;
(v) fall on an empty region with no overlap—this is a true FP, but it is rare (\textbf{0.2} per frame in our statistics).
In practice, (i) and (ii) dominate, while (v) is negligible; consequently, \emph{FP/frame} stays near baseline as \emph{FN/frame} drops (Appendix~\Cref{fig:fp_analysis}).}

\subsection{Ablation Studies}
\label{sec:4.5}

To better understand the key design choices in \algname, we conduct a series of ablation studies. Specifically, we investigate: \textit{(i)} the contribution of core alignment components such as similarity and localization losses, \textit{(ii)} the effect of addressing instance ambiguity through a one-to-$N$ mapping strategy, \textit{(iii)} the capability of visual prompt alignment in achieving instance-level object differentiation, and \textit{(iv)} the dynamic behavior of the visual prompt buffer during online inference. These studies provide deeper insights into the effectiveness and robustness of \algname in leveraging visual prompts for online 3D detection.

\myparagraph{Effect of Components in Visual Prompt Alignment.}
\begin{table}
    \centering
    \caption{\textbf{Effect of Visual Prompt Alignment.} We select \algname-MonoDETR for this experiment.}
    \label{tab:alignhead}
    \small
    \setlength{\tabcolsep}{2.5mm}
    \begin{tabular}{c|c|c|c}
    \toprule
         \makecell[l]{Sim. \\ Loss} & \makecell[l]{Loc. \\ Loss} & mAP (\%) & EDS (\%) \\
        \midrule
        - & - & 32.6 & 44.2 \\
        $\surd$ & - & 39.4 & 48.5 \\
        $\surd$ & $\surd$ & \bf 43.3 & \bf 52.3 \\
    \bottomrule
    \end{tabular}
\end{table}
Visual prompt alignment aims to localize objects via visual prompts in input images. We now evaluate its components with \algname-MonoDETR.
\Cref{tab:alignhead} presents experiments to verify the core components of this alignment, including similarity loss (Focal and Dice loss) and localization loss (supervising visual prompt localization, $\mathcal{X}_v$).
While the alignment can be implicitly learned by attention mechanisms in the transformer decoder, incorporating explicit similarity supervision brings improvements of 6.8\% mAP and 4.3\% EDS.
Further utilizing the position loss and one-to-$N$ mapping (to address instance ambiguity) boosts the performance to 43.3\% mAP and 52.3\% EDS.
These results prove this alignment operation is a critical component enabling the \algname detectors to effectively detect target objects via visual prompts.
\begin{figure*}[ht]
    \centering
     \begin{minipage}{0.5\textwidth}
        \centering
      \includegraphics[width=0.9\linewidth]{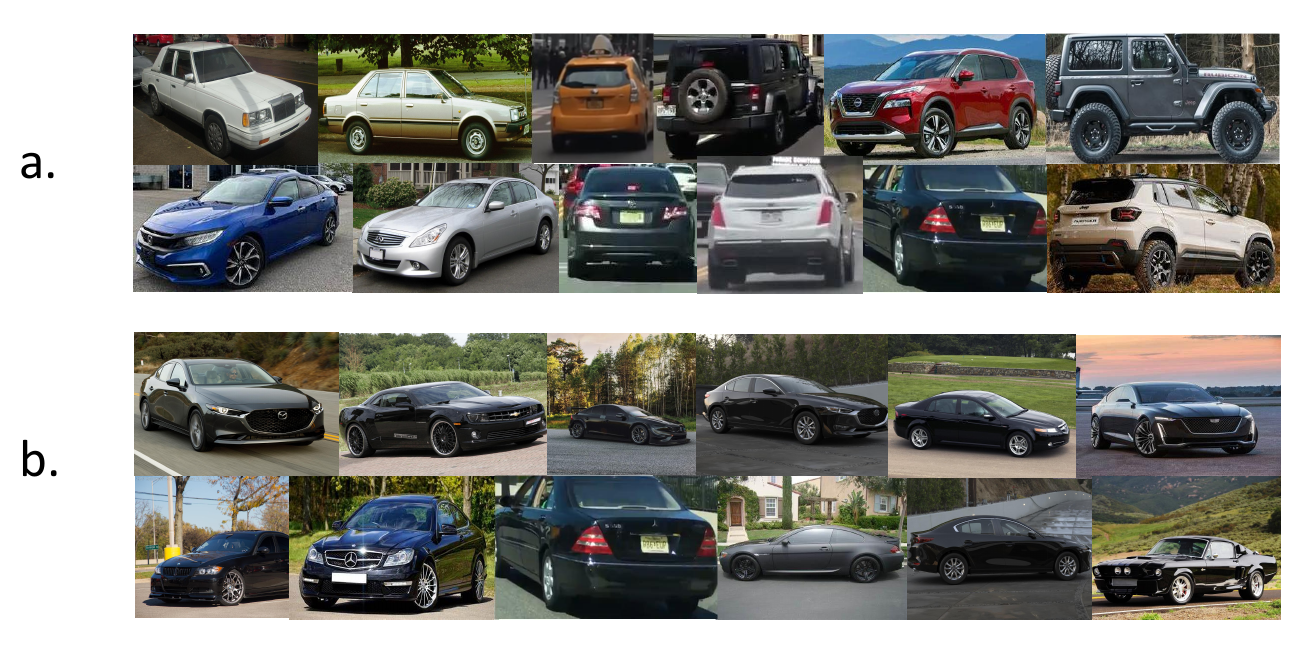} 
        \caption{Visual prompt showcase for \Cref{tab:all_black_car}. }
        \label{fig:web_prompts_black}
    \end{minipage}\hfill
    \begin{minipage}{0.5\textwidth}
        \centering
        \captionof{table}{\textbf{Effect of visual prompt alignment on instance awareness.} Group (a.) resembles ``car'' objects in nuScenes validation set with diverse types and colors; while group (b.) maintains black sedans only. The performance gap demonstrates that alignment learns the instance awareness.}
        \label{tab:all_black_car}
        \begin{tabular}{c|cc}
        \toprule
          Group & mAP (\%) $\uparrow$ & EDS (\%) $\uparrow$ \\ 
        \midrule
        \midrule
           a. &  \bf 20.9   & \bf 45.3   \\
        
         b. & 13.8    &   42.2 \\

        \bottomrule
    \end{tabular}
    \end{minipage}
      
\end{figure*}

\myparagraph{Instance Ambiguity in Visual Prompt Alignment.}
\begin{table}[t]
    \centering
    \caption{\textbf{Effect of the number of predicted positions $N$ of visual prompt alignment.} We select \algname-MonoDETR for this experiment.}
    \label{tab:alignhead_ambiguity}
    \small
    \setlength{\tabcolsep}{3.mm}
    \begin{tabular}{c|c|c}
    \toprule
        \makecell[l]{No. of Position \\ Predictions} & mAP (\%) $\uparrow$ & Recall (\%) $\uparrow$ \\
        \midrule
        1 & 39.9 & 62.1 \\
        4 & \bf 43.3 & 69.1 \\
        8 & 43.0 & \bf 69.4 \\
    \bottomrule
    \end{tabular}
\end{table}
To solve the instance ambiguity issue, we propose to regress $N$ positions of each visual prompt when performing the visual prompt alignment. This retrieves all objects with similar visual contents.
In this study, we validate the effectiveness of this design by conducting ablation studies on the number of $N$. 
As listed in \Cref{tab:alignhead_ambiguity}, when the number of position predictions $N$ equals 1, which means a one-to-one mapping for each visual prompt, the mAP and recall rate are 39.9\% and 62.1\%, respectively. 
Then, if we increase the $N$ to 4, effectively a one-to-four mapping, we obtain an mAP of  43.3\% and a recall rate of 69.1\%. 
This represents a 7\% improvement in the recall rate, demonstrating that instance ambiguity is an important challenge in visual prompt-based detection, and the proposed one-to-$N$ mapping solution effectively addresses this issue.

\myparagraph{Effect of Visual Prompt Alignment on Instance Awareness.}
Despite demonstrating that visual prompt alignment can enhance system performance, we remain uncertain whether the alignment can distinguish different objects based on visual prompts for instance-level matching. 
\begin{figure*}[t]
	\centering
        \includegraphics[width=1.0\linewidth]{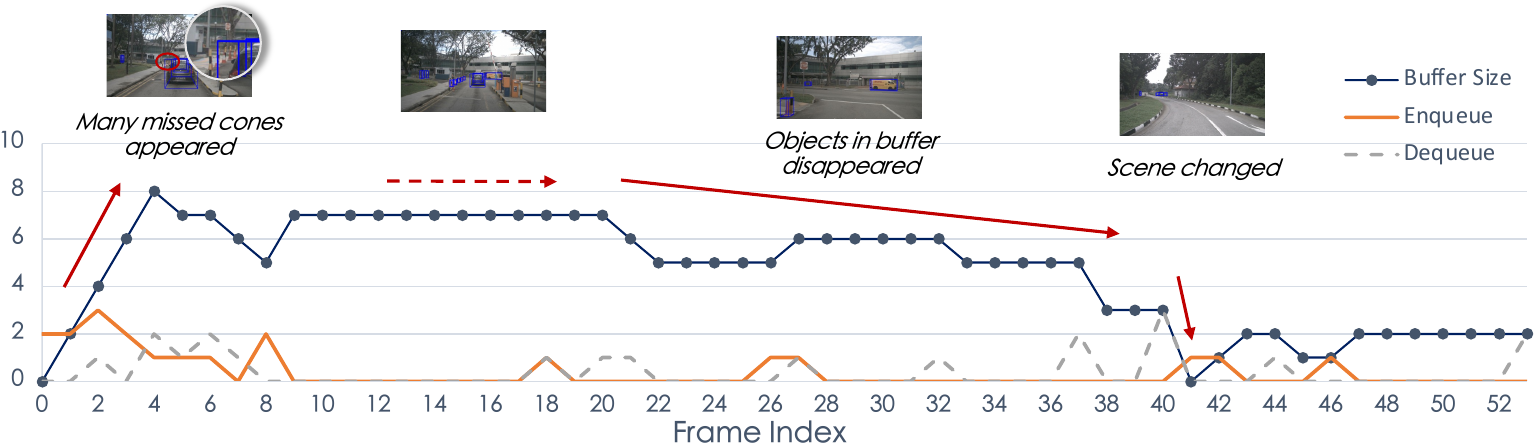}
	\vspace{-0.2in}
	\caption{\textbf{Size of visual prompt buffer during the video stream.} Visual prompt buffer stores the missed objects during online inference to rectify test-time errors of deployed 3D detectors. It can adaptively manage the stored prompts and thus maintain the balance between latency and accuracy.}
	\vspace{-0.15in}
	\label{fig:supp_exp_dynamicbuffer}
\end{figure*}
To investigate, we conduct an additional experiment, based on a similar setting with \Cref{tab:web_prompts} (\algname-MonoDETR trained on 20\% vehicle annotations and freeze the prompt buffer with predefined web prompts when inference.), but fix the visual prompt buffer with images of black sedans solely (\Cref{fig:web_prompts_black}).
As shown in \Cref{tab:all_black_car}, as ``Car'' objects in nuScenes contain various types and colors, solely using black sedans as visual prompts leads to a 7.1\% mAP drop.
This underscores that the alignment operation effectively differentiates objects based on visual prompts, achieving instance-level matching and detection.

\myparagraph{Statistics of Visual Prompt Buffer during Online Inference.}
We design the visual prompt buffer to store missed objects during inference with the video stream and introduce a ``dequeue'' mechanism to prevent the buffer from growing indefinitely. 
In this ablation study, we analyze the dynamic buffer size, as well as the number of enqueued and dequeued in each frame, to illustrate the behavior of the visual prompt buffer during the online operation of \algname.
As shown in \Cref{fig:supp_exp_dynamicbuffer}, the visual prompt buffer exhibits three distinct behaviors during online inference in each nuScenes video clip: increasing, steady, and decreasing. 
In initial frames, many traffic cones are queued into the buffer due to the poor performance of deployed offline 3D detectors on cone objects. 
The buffer size thus grows quickly in initial frames to store visual prompts of missed objects for online rectification (Frames \#0 to \#4). 
The buffer size then stabilizes as the online detector consistently detects and tracks all objects of interest (Frames \#4 to \#20).
Further, as the ego vehicle drives out of the scene, many previously enqueued objects no longer exist and are thus removed from the buffer automatically (Frames \#20 to \#40). The buffer finally becomes empty as the scene changes.
This demonstrates the effectiveness of the visual prompt buffer, which consistently stores missed objects during online inference and corrects online errors.
This dynamic behavior, exhibiting increasing, steady, and decreasing phases, highlights itself to manage stored visual prompts for robust 3D object detection performance throughout the online inference with the balance between latency and accuracy.
\mrev{We also report several compact, deployment-oriented statistics over the full validation stream on \algname-BEVFormer:
(i) the \emph{largest} buffer size observed over the entire stream is only \textbf{65}, with an \emph{average} of \textbf{3.9};
(ii) after an object leaves the scene, only \textbf{10.3}\% of its prompts remain in the buffer;
(iii) only \textbf{17.2}\% of those left objects in (ii) produce erroneous final box after standard post-processing (confidence filtering + NMS).
Together, these results indicate that bounded buffer growth, timely removal of outdated prompts, and negligible impact on final predictions—consistent with the dynamics in \Cref{fig:supp_exp_dynamicbuffer}.}


\section{Conclusion}
\label{sec:sec5}

In this paper, we introduce the \algname system. It equips existing 3D detectors with the ability to make test-time error corrections. The core component is the \modelname, which enables offline-trained 3D detectors with the ability to leverage visual prompts for continuously detecting and tracking previously missing 3D objects. By updating the visual prompt buffer, \algname system enables continuous error rectification online without any training. To conclude, \algname provides a more reliable online 3D perception system, allowing seamless transfer of offline-trained 3D detectors to new autonomous driving deployments. We hope this work will inspire the development of online correction systems.

\myparagraph{Limitations and Future Work.}
As the first work on online 3D detection, this study has several limitations, which might open up many potential research directions. 

For performance, the generalization ability is constrained by the model size and data volume, and future work will aim to enhance this by scaling up both the data and parameters. 

For model design, while allowing auxiliary information to provide reliability, it introduces potential risks.
Requiring frequent feedback, especially human feedback, may be challenging in current L3 autonomous driving paradigms. Though reducing the frequency of feedback can alleviate this pressure without compromising performance, there is room for improvement. Future work could integrate robust tracking methods or leverage advanced hardware, such as eye-tracking, to simplify the feedback collection.

For objectives, this work focuses on addressing missed detections. It does not tackle the redundancy issues. Future work could explore addressing false positive issues through \algname system. Additionally, to enable zero-shot detection, this work adopts a class-agnostic approach. In the future, methods combining language models could be explored, ensuring both low latency and open-world category detection ability. Finally, this study only explores error correction within the visual domain; extending this approach to other sensors, such as LiDAR, will be an area for further investigation.

\section*{Acknowledgement}
This work was supported by National Key R\&D Program of China (2022ZD0160104), NSFC (62206172), and Shanghai Committee of Science and Technology (23YF1462000).
This work is in part supported by the JC STEM Lab of Autonomous Intelligent Systems funded by The Hong Kong Jockey Club Charities Trust.
We thank team members from OpenDriveLab for valuable feedback along the project. 
Special thanks to Chengen Xie for providing data support, and Jiahui Fu for sharing insights on MV2D. We also sincerely thank Qingsong Yao for providing writing guidance and suggestions on experimental design.

\bibliography{bibliography}
\bibliographystyle{ieeetr}

\vfill
\clearpage

{\appendices
\section{Discussions}
\label{sec:discussion}

For a better understanding of our work, we supplement intuitive questions that one might raise. Note that the following list does \textit{not} indicate whether the manuscript was submitted to a previous venue or not.

\medskip
\myparagraph{Q1:} \textit{What is the relationship between the online 3D detection and the offline data-loop progress?}

We emphasize, in this paper, we do not propose the online system to replace the traditional offline data loop. 
As illustrated in~\Cref{fig:teaser_visual_prompt}, these two systems address different aspects of autonomous driving. 
The offline system remains crucial for enhancing the capabilities of the base perception model through development; while our online \algname system further enables the deployed frozen model in vehicles to promptly rectify dangerous driving behaviors caused by unrecognized objects on the road. 
With improved offline-trained detectors, the \algname can effectively correct more online errors during test-time inference without re-training, as detailed in~\Cref{tab:main}.

\medskip
\myparagraph{Q2:} \textit{What are the main technical novelty and advantages of the proposed \algname system over previous
instruction-based 3D detectors?}

The advantages lie in the design of visual prompts, the visual descriptions of target objects with diverse sources, styles, poses, and timestamps. While existing instruction-based 3D detection methods typically utilize text, boxes, or clicks as prompts.

Compared to text prompts, visual prompts provide a more natural and accurate description of the target object. In contrast, verbal descriptions can be ambiguous to convey instance-level features, leading to an inaccurate understanding of the missing objects. Second, text promptable models are often combined with LLMs with high latency and are thus unavailable for autonomous driving deployment, as discussed in~\Cref{sec:4.4}.

Box and point prompts are less convenient than visual prompts when dealing with stream data. If missing occurs, these single-frame prompts require feedback at every frame, which is unfeasible in real-world applications.  Compared to box and point prompts, visual prompts are robust across different scenes and timestamps, one single-frame visual prompt is enough for detecting and tracking in later frames.
Furthermore, visual prompts enable 3D detection with pre-defined visual descriptions of target objects, regardless of the sources, styles, poses, etc, as discussed in~\Cref{sec:4.3}.

The introduction of novel visual prompts enables real-time, accurate, and continuous error correction of streaming inputs.

\medskip
\myparagraph{Q3:} \textit{What is the relationship between the \algname system and existing 3D perception tasks?}

The \algname system relates to several 3D perception tasks, including 3D object detection, zero/few-shot detection, domain adaptation, single object tracking, and continual learning.

Compared to standard 3D object detection, the primary focus of the \algname system is on enabling instant online error correction rather than optimizing the offline detection performance of the base 3D detector. In contrast to traditional few-shot, one-shot, or domain adaptation approaches, the \algname system does not require 3D annotations for new objects or any model retraining, yet can still provide reasonable 3D bounding box estimates for out-of-distribution objects. 
Relative to image-based single object tracking, \algname is not merely limited to generating 2D detection outputs and assuming the object query as the initial frame of an ongoing video. Instead, it is capable of performing 3D tracking based on visual descriptions of target objects from any scene or timestamp, leveraging a diverse set of visual prompts.

In summary, the \algname system represents a more flexible and comprehensive 3D object detection framework, combining the strengths of zero-shot detection, handling out-of-distribution objects, and utilizing diverse visual prompts beyond the current scene context.

\medskip
\myparagraph{Q4:} \textit{What are potential applications and future directions of \algname?}

We believe that, visual prompts, as the core design element of the \algname system, represent a more natural and intuitive query modality for the image domain. This approach has significant research potential and application prospects in the field of 3D perception and beyond.

For example, visual prompts enable rapid customization of the tracking targets, beyond the pre-defined object classes. Second, the visual prompt-based framework facilitates online continual learning for 3D perception systems, adapting to evolving environments. Then, visual prompts can be applied in the V2X domain to enable swift error rectification across diverse operational scenarios. Visual prompts can also be deployed to assist in the auto-labeling process of target objects. Furthermore, by combining visual prompts with natural language prompts, we can obtain more precise descriptions and behavioral control for online perception systems.

The diverse applications outlined above demonstrate the promise of visual prompts as a versatile approach. As showcased in this work, the visual prompt-based framework opens up new possibilities for online perception systems, not only in autonomous driving but also in a broader range of domains.

\section{Additional Explanations}
\label{sec:supp_explanation}
\mrev{This section provides visual explanations and additional tables supporting analyses in the main paper.}

\begin{figure}
  \centering
    \includegraphics[width=0.5\textwidth]{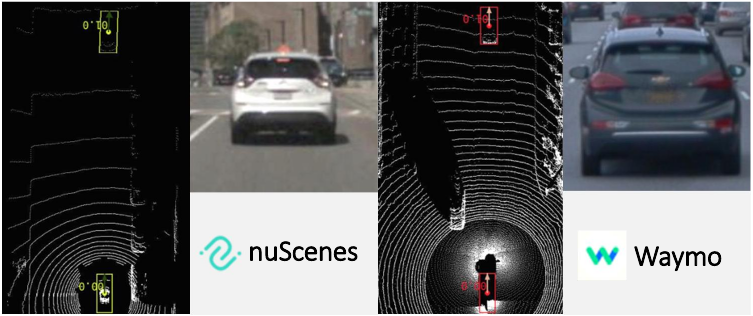}
  \caption{\mrev{\textbf{Intrinsic mismatch across datasets.} Equal-size image patches around cars at comparable 3D distance/shape from nuScenes (left) and Waymo (right). Despite similar geometry, the apparent 2D scale differs markedly due to different focal lengths and camera rigs, inducing depth bias and explaining why naive cross-dataset transfer of vision-only 3D detectors degrades severely.
  \textit{Adapted from} \cite{zheng2023cross}}.}
  \label{fig:cross_dataset}
\end{figure}
\begin{figure}
  \centering
    \includegraphics[width=0.46\textwidth]{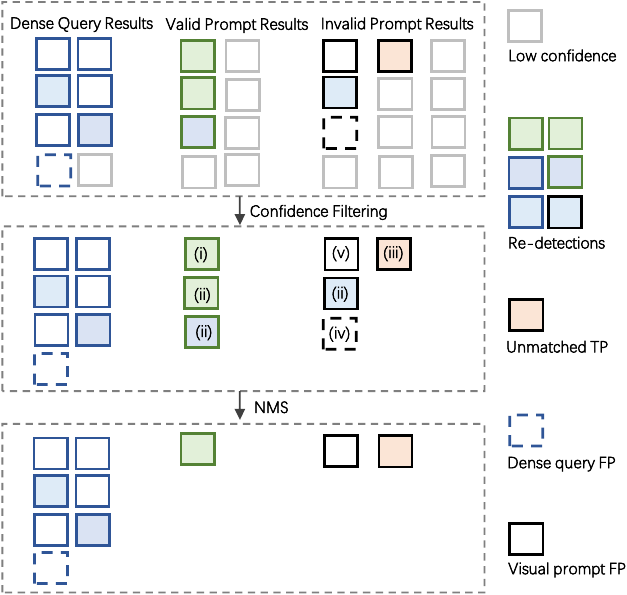}
  \caption{\mrev{\textbf{Outcomes of visual prompts and their FP impact.} Final predictions consist of dense-query results and visual-prompt results. Visual prompts produce few above-threshold candidates, and those candidates fall into five cases (i–v). After confidence filtering and NMS, only a small portion contributes to false positives (v), explaining why \emph{FP/frame} remains near baseline.}}
  \label{fig:fp_analysis}
\end{figure}

\subsection{\mrev{Cross-dataset Transfer.}}
\label{sec:supp_cross_dataset_analysis}
\mrev{\Cref{fig:cross_dataset} visualizes the intrinsic mismatch across datasets, adapted from \cite{zheng2023cross}.
Equal-size image patches from nuScenes (left) and Waymo (right) show that objects at comparable 3D distance and shape appear at different 2D scales due to camera focal-length and rig disparities.
This visualization supports the cross-dataset transfer analysis in~\Cref{sec:4.4}, explaining why naive cross-dataset transfer of vision-only 3D detectors leads to depth bias and severe degradation.}

\subsection{\mrev{False-Positive Impact.}}
\label{sec:supp_fp_analysis}
\mrev{\Cref{fig:fp_analysis} illustrates five cases of visual prompt results.
After confidence filtering and NMS, valid prompts mainly produce correct detections or re-detections, while invalid prompts rarely survive.
The remaining above-threshold cases (i–v) explain why TTC reduces missed detections without materially increasing false positives, as discussed in FP Analysis in~\Cref{sec:4.4}.
}
\mrev{To corroborate the deployment-relevant regime, Table~\ref{tab:fp_fn_tau03} reports \emph{FP/frame} and \emph{FN/frame} at a confidence threshold of $\tau{=}0.3$; changes in \emph{FP/frame} are negligible and not statistically significant across backbones.}

\begin{table}[t]
\centering
\small
\setlength{\tabcolsep}{6pt}
\caption{\mrev{\textbf{False-alarm metrics at a deployment-oriented confidence threshold ($\tau{=}0.3$).}
}}
\label{tab:fp_fn_tau03}
\begin{tabular}{lcc}
\toprule
\textbf{Method} & \textbf{FP/frame@0.3} $\downarrow$ & \textbf{FN/frame@0.3} $\downarrow$ \\
\midrule
MonoDETR & 1.5 & 2.5 \\
\cellcolor{LightCyan}TTC-MonoDETR & \cellcolor{LightCyan}1.5 & \cellcolor{LightCyan}2.1 \\
\midrule
MV2D & 1.3 & 2.9 \\
\cellcolor{LightCyan}TTC-MV2D & \cellcolor{LightCyan}1.3 & \cellcolor{LightCyan}2.2 \\
\midrule
Sparse4Dv2 & 1.1 & 3.0 \\
\cellcolor{LightCyan}TTC-Sparse4Dv2 & \cellcolor{LightCyan}1.1 & \cellcolor{LightCyan}2.2 \\
\midrule
BEVFormer & 1.2 & 2.9 \\
\cellcolor{LightCyan}TTC-BEVFormer & \cellcolor{LightCyan}1.2 & \cellcolor{LightCyan}2.1 \\
\midrule
BEVFormerV2-t8 & 1.2 & 2.9 \\
\cellcolor{LightCyan}TTC-BEVFormerV2-t8 & \cellcolor{LightCyan}1.2 & \cellcolor{LightCyan}2.1 \\
\midrule
RayDN & 1.3 & 2.9 \\
\cellcolor{LightCyan}TTC-RayDN & \cellcolor{LightCyan}1.3 & \cellcolor{LightCyan}2.1 \\
\midrule
StreamPETR & 1.3 & 2.8 \\
\cellcolor{LightCyan}TTC-StreamPETR & \cellcolor{LightCyan}1.3 & \cellcolor{LightCyan}2.3 \\
\bottomrule
\end{tabular}
\end{table}

\section{Qualitative Results}
\label{sec:supp_vis}
We provide extensive visualizations to demonstrate the versatility of the \algname system across diverse scenarios:
\begin{itemize}

\item In \Cref{sec:E.1} and \Cref{sec:E.2}, we fix the prompt buffer with visual prompts of either labeled or novel, unlabeled 3D objects from the nuScenes dataset to detect targets in the nuScenes images.

\item In \Cref{sec:E.3}, we test the performance with prompt buffer containing visual prompts in styles differing from the training distribution, such as Lego.

\item In \Cref{sec:E.4}, we visualize the similarity maps on out-of-domain images, including YouTube driving videos and Internet-sourced visual prompts, demonstrating the generalization of the visual prompt alignment and the effectiveness of our \algname in reducing driving risks in non-standard scenarios.

\end{itemize}

For all visualizations, the fixed prompt buffer is shown in the first row.
These comprehensive evaluations highlight the versatility of the \algname system in leveraging diverse visual prompts for 3D detection.
All results are conducted with \algname-MonoDETR.

\subsection{In-domain Visual Prompts on nuScenes ``Seen'' Objects.}
\label{sec:E.1}

This visualization focuses on the in-domain detection performance of the \algname system on the nuScenes dataset. We utilize visual prompts from \textit{labeled} objects in the nuScenes dataset, and demonstrate the system's ability to effectively detect and track these target objects across different frames, as shown in \Cref{fig:app-c1-1} and \Cref{fig:app-c1-2}.

\subsection{In-domain Visual Prompts on nuScenes ``Unseen'' Objects}
\label{sec:E.2}
This visualization focuses on the \algname's ability to handle visual prompts of objects \textit{not labeled} in the nuScenes dataset. As shown in \Cref{fig:app-c2-1}, \Cref{fig:app-c2-2}, and \Cref{fig:app-c2-3}, our method demonstrates its potential to detect and track novel, out-of-distribution objects with these unseen visual prompts.

\subsection{Out-domain Visual Prompts on nuScenes ``Seen'' Objects}
\label{sec:E.3}
This visualization focuses on the \algname's performance with visual prompts in \textit{styles different from the training distribution}. As shown in \Cref{fig:app-c3-1} and \Cref{fig:app-c3-2}, the model can effectively detect target objects using visual prompts in various views and styles that diverge from the original training data.

\subsection{Out-domain Visual Prompts on Real-world Examples}
\label{sec:E.4}

This visualization examines the generalization and robustness of \algname detectors in aligning visual prompts with the corresponding target objects in the input video stream. We select challenging driving scenarios involving unexpected animals running into the path of the ego vehicle. This is aimed at demonstrating the \algname system's capability in reducing online driving risks in such non-standard situations.

As shown in \Cref{fig:app-c5-1} and \Cref{fig:app-c5-2}, our method can effectively localize non-expected animals with higher responses in the regions where animals located, even though it was trained solely on the nuScenes dataset. \Cref{fig:app-c5-3}, \Cref{fig:app-c5-4}, \Cref{fig:app-c5-5}, \Cref{fig:app-c5-6} further present examples of detection in a real-world scenario containing both vehicles and animals. Our method can detect all objects with their visual prompts simultaneously.

\subsection{License of Assets}
\label{sec:supp-license}

The adopted nuScenes dataset~\cite{nuscenes2019} is distributed under a CC BY-NC-SA 4.0 license. We implement the model based on mmDet3D codebase~\cite{mmdet3d2020}, which is released under the Apache 2.0 license.

We will publicly share our code and models upon acceptance under Apache License 2.0.

\clearpage
\begin{wrapfigure}{r}{\textwidth}
  \centering
    \includegraphics[width=0.7\textwidth]{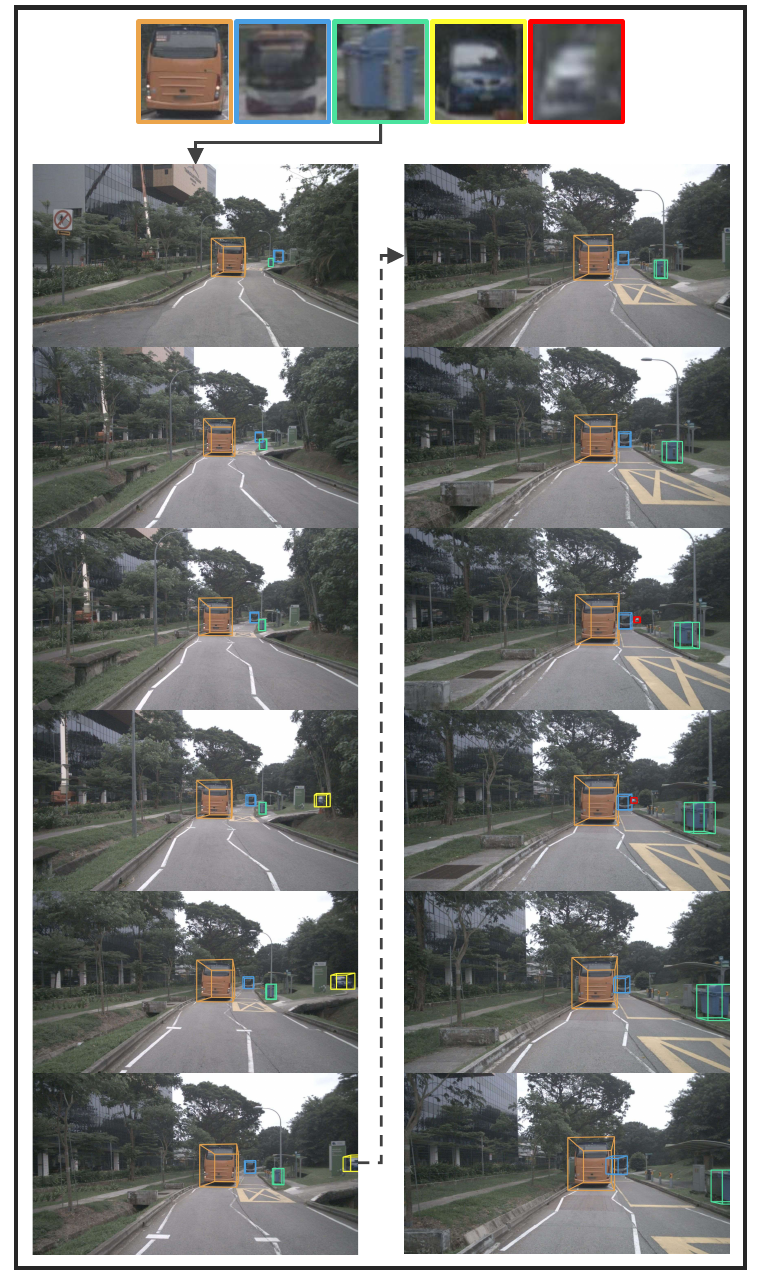}
  \vspace{-0.05in}
  \caption{\textbf{Visualizations on nuScenes scenarios with in-domain visual prompts of \textit{labeled} objects.} \algname system enables continuous 3D detection and tracking based on visual prompts. The images in the first row indicate the visual prompts in prompt buffer, and images in other rows represent 3D detection results prompted by the corresponding visual prompts. Different identities are indicated with different colors.}
  \label{fig:app-c1-1}
\end{wrapfigure}

\clearpage
\begin{wrapfigure}{r}{\textwidth}
  \centering
    \includegraphics[width=0.7\textwidth]{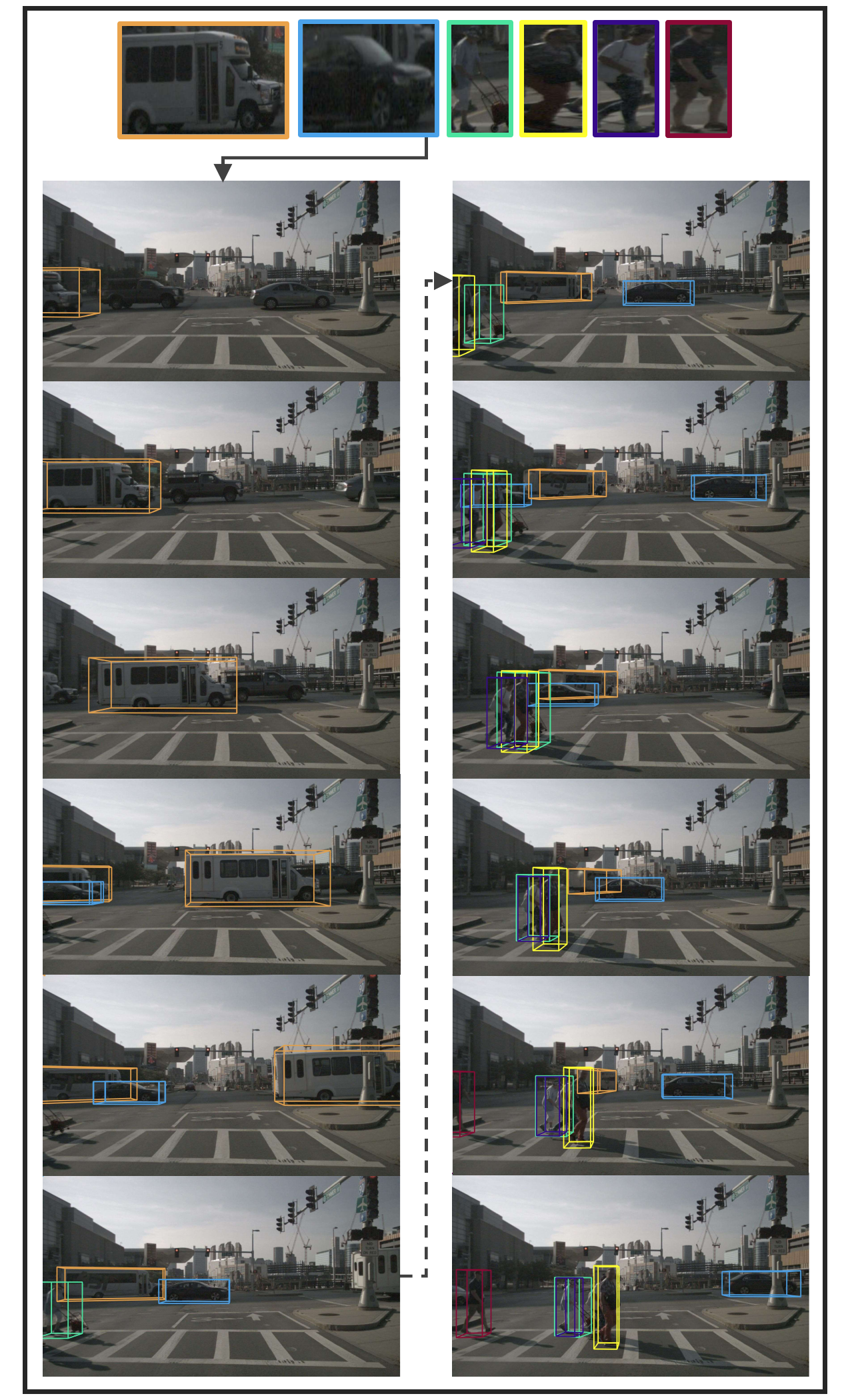}
  \caption{\textbf{Visualizations on nuScenes scenarios with in-domain visual prompts of \textit{labeled} objects.} \algname system enables continuous 3D detection and tracking based on visual prompts. The images in the first row indicate the visual prompts in prompt buffer, and images in other rows represent 3D detection results prompted by the corresponding visual prompts. Different identities are indicated with different colors.}
  \vspace{-0.05in}
  \label{fig:app-c1-2}
\end{wrapfigure}

\clearpage
\begin{wrapfigure}{r}{\textwidth}
  \centering
    \includegraphics[width=0.9\textwidth]{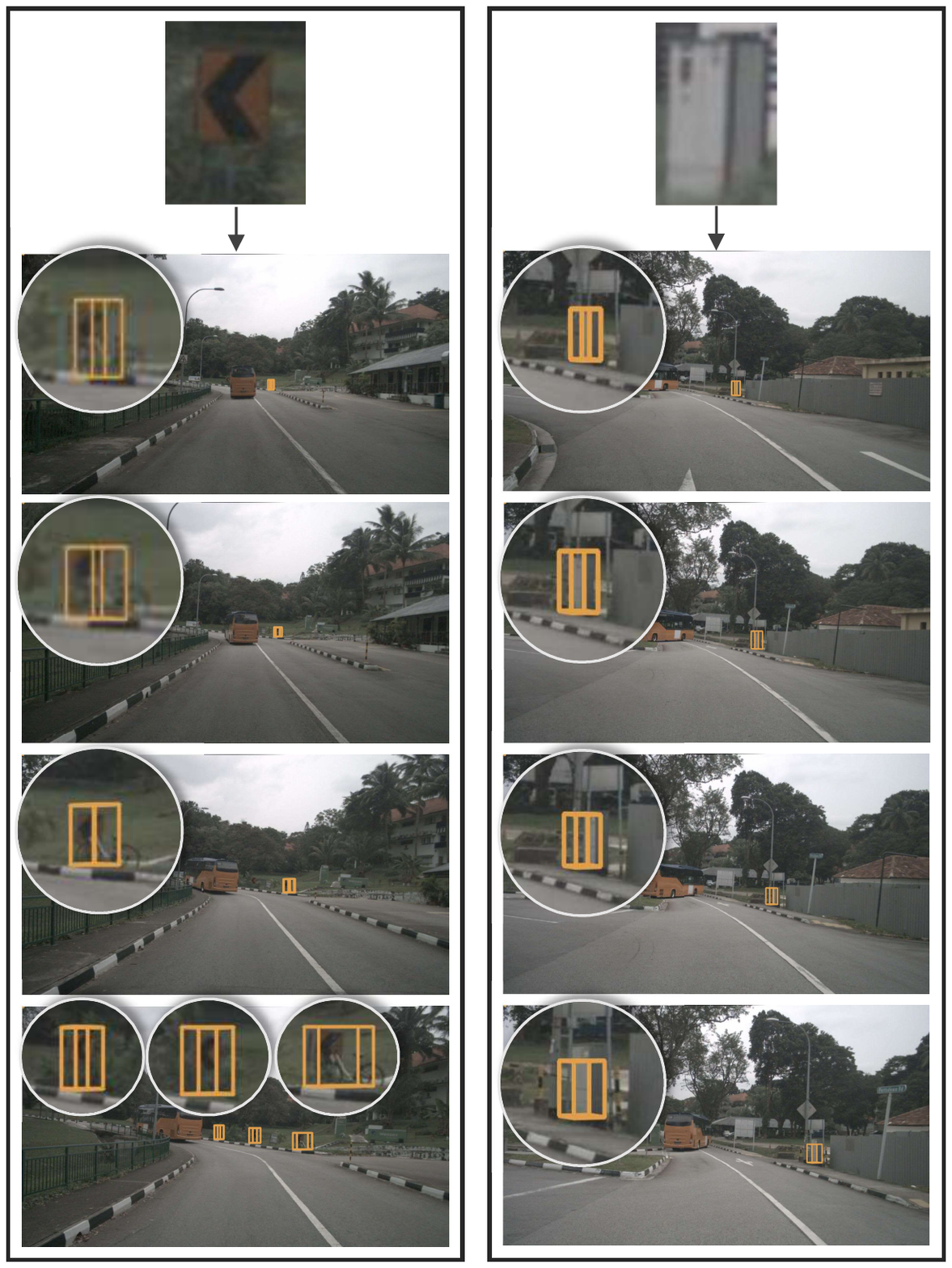}
  \vspace{-0.1in}
  \caption{\textbf{Visualizations on nuScenes scenarios with in-domain visual prompts of \textit{un-labeled} objects.} \algname system enables 3D detection and tracking of ``novel'' objects unseen during training. The image in the first row indicates the visual prompt in the prompt buffer, and images in other rows represent 3D detection results prompted by the corresponding visual prompts. Interestingly, with the one-to-N mapping mechanism of the visual prompt alignment, \algname system can detect multiple objects with similar visual descriptions to the visual prompt simultaneously.}
  \vspace{-0.05in}
  \label{fig:app-c2-1}
\end{wrapfigure}

\clearpage
\begin{wrapfigure}{r}{\textwidth}
  \centering
    \includegraphics[width=0.9\textwidth]{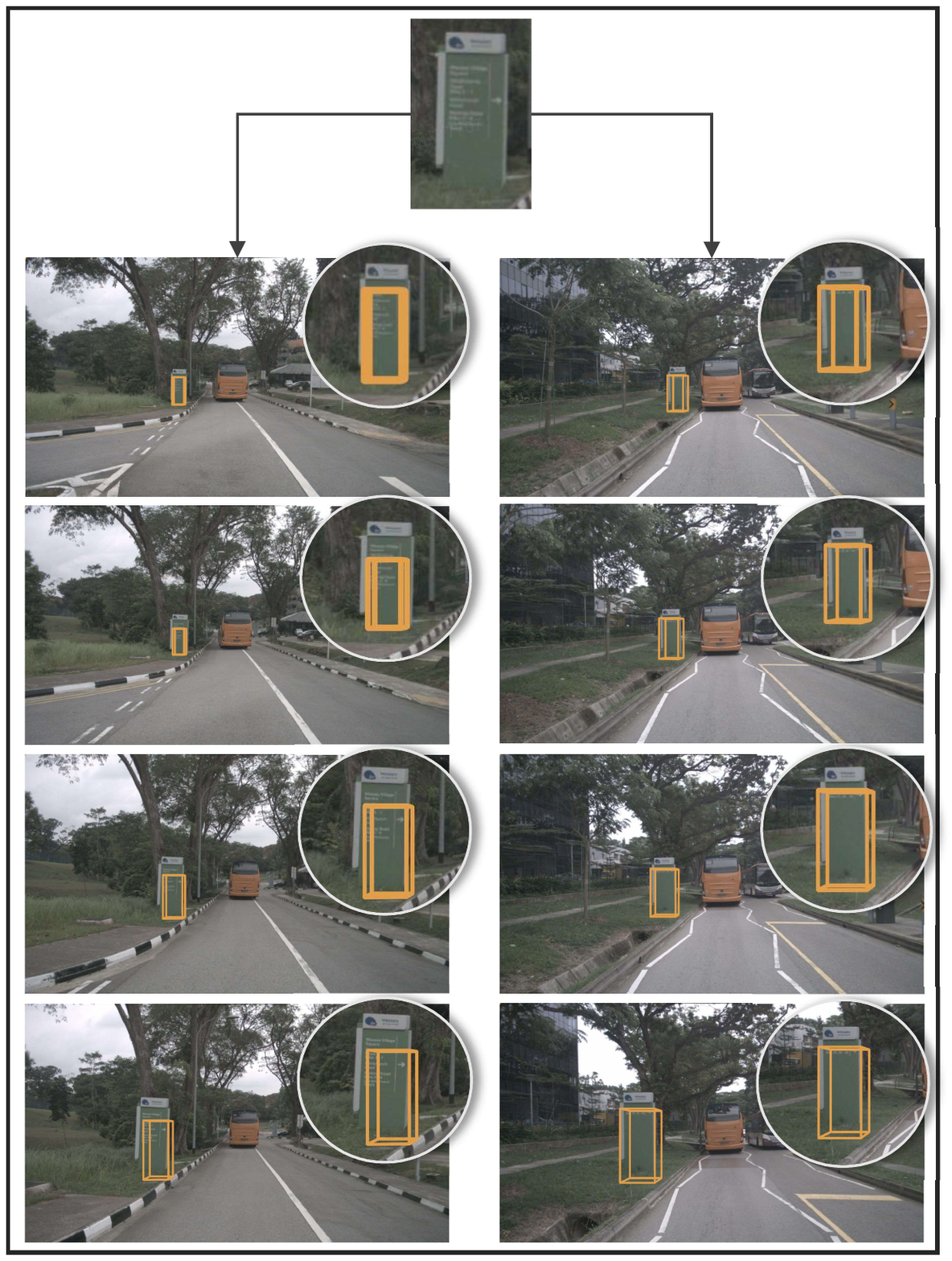}
  \vspace{-0.1in}
  \caption{\textbf{Visualizations on nuScenes scenarios with in-domain visual prompts of \textit{un-labeled} objects.} \algname system enables 3D detection and tracking of ``novel'' objects unseen during training. The image in the first row indicates the visual prompt in the prompt buffer, and images in other rows represent 3D detection results prompted by the corresponding visual prompts. }
  \vspace{-0.05in}
  \label{fig:app-c2-2}
\end{wrapfigure}

\clearpage
\begin{wrapfigure}{r}{\textwidth}
  \centering
    \includegraphics[width=0.9\textwidth]{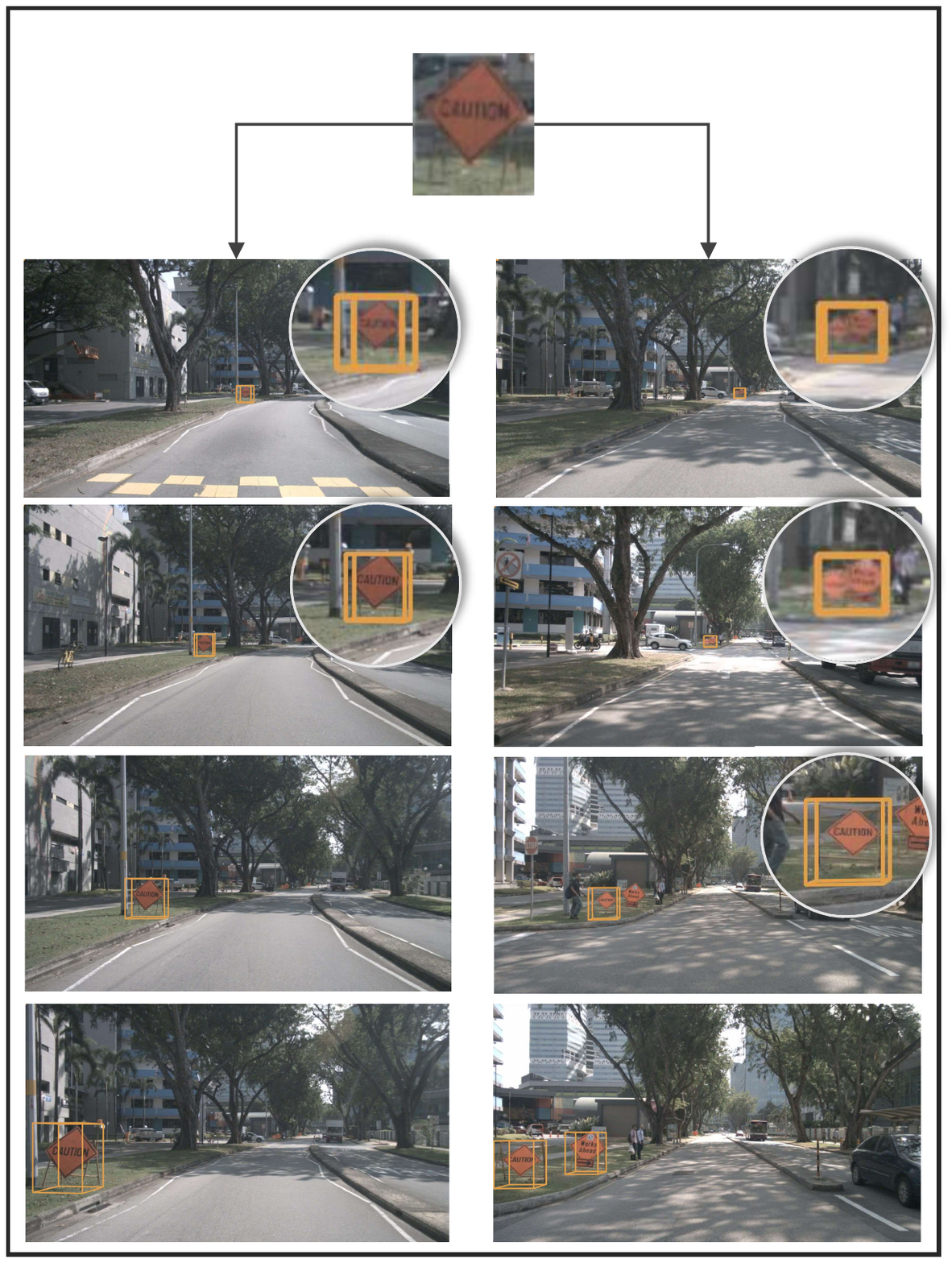}
  \vspace{-0.1in}
 \caption{\textbf{Visualizations on nuScenes scenarios with in-domain visual prompts of \textit{un-labeled} objects.} \algname system enables 3D detection and tracking of ``novel'' objects unseen during training. The image in the first row indicates the visual prompt in the prompt buffer, and images in other rows represent 3D detection results prompted by the corresponding visual prompts. Interestingly, with the one-to-N mapping mechanism of the visual prompt alignment, \algname system can detect multiple objects with similar visual descriptions to the visual prompt simultaneously.}
  \vspace{-0.05in}
  \label{fig:app-c2-3}
\end{wrapfigure}
\clearpage
\begin{wrapfigure}{r}{\textwidth}
  \centering
    \includegraphics[width=0.9\textwidth]{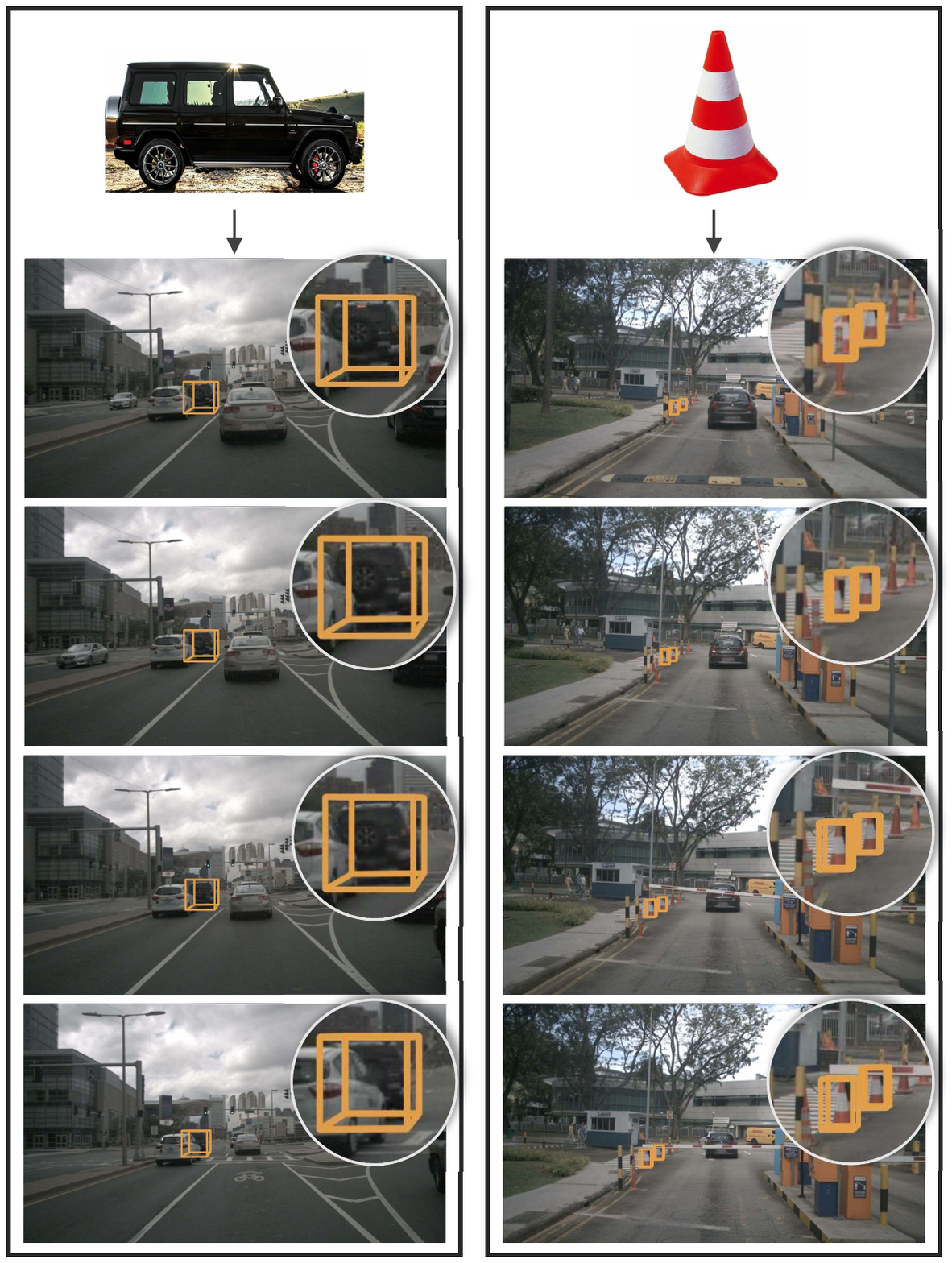}
  \vspace{-0.1in}
  \caption{\textbf{Visualizations on nuScenes scenarios with \textit{out-domain} visual prompts of \textit{labeled} objects.}
  \algname system can perform 3D detection and tracking via visual prompts with arbitrary styles (\textbf{imagery style}). The image in the first row indicates the visual prompt in the prompt buffer, and images in other rows represent 3D detection results prompted by the corresponding visual prompts. With the one-to-N mapping mechanism of the visual prompt alignment, \algname detectors can detect multiple objects with similar visual descriptions to the visual prompt at the same time.}
  \vspace{-0.05in}
  \label{fig:app-c3-1}
\end{wrapfigure}

\clearpage
\begin{wrapfigure}{r}{\textwidth}
  \centering
    \includegraphics[width=0.9\textwidth]{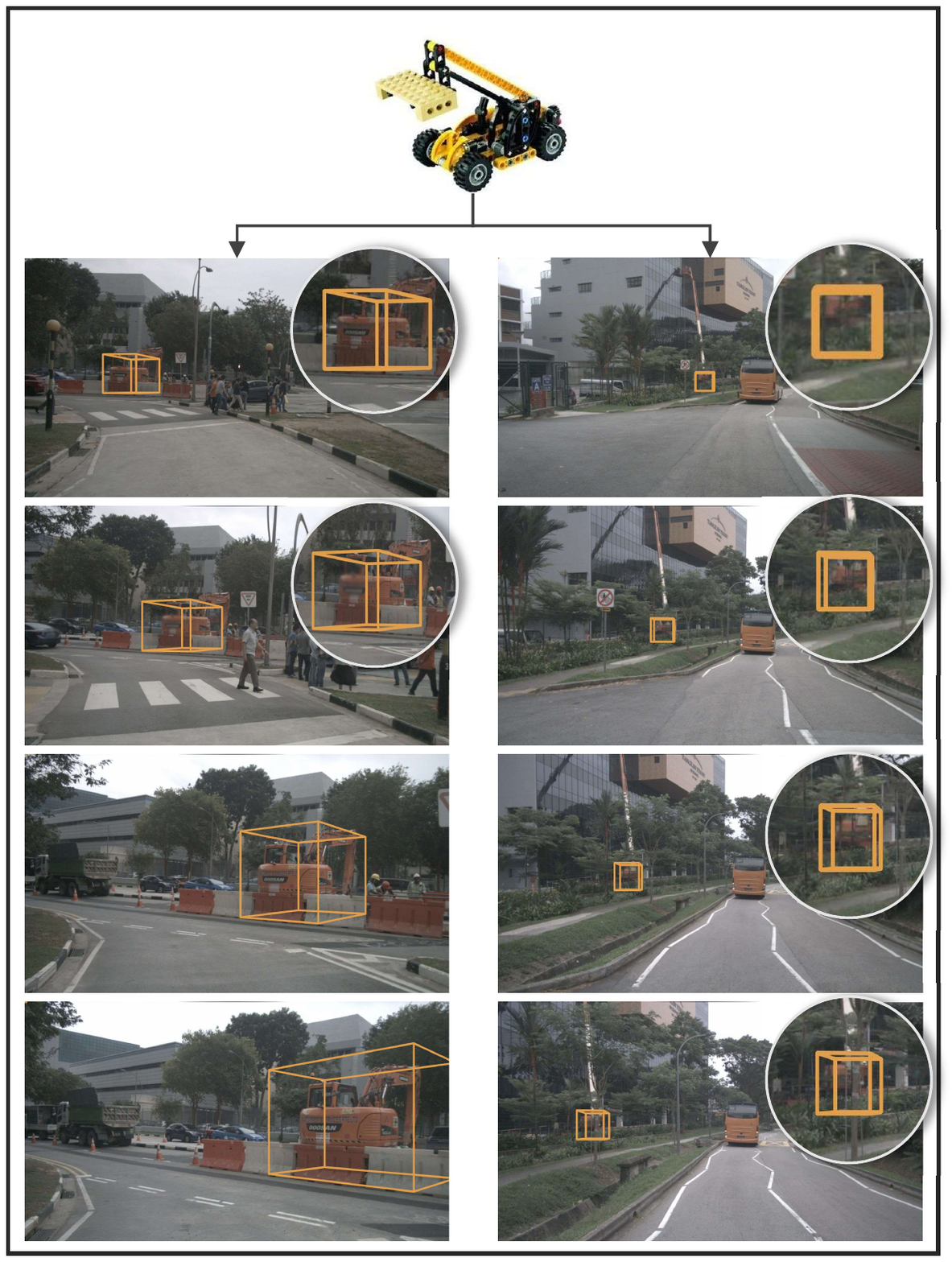}
  \vspace{-0.1in}
  \caption{\textbf{Visualizations on nuScenes scenarios with \textit{out-domain} visual prompts of \textit{labeled} objects.}
  \algname system can perform 3D detection and tracking via visual prompts with arbitrary styles (\textbf{Lego style}). The image in the first row indicates the visual prompt in the prompt buffer, and images in other rows represent 3D detection results prompted by the corresponding visual prompts. With the one-to-N mapping mechanism of the visual prompt alignment, \algname detectors can detect multiple objects with similar visual descriptions to the visual prompt at the same time.}
  
  \vspace{-0.05in}
  \label{fig:app-c3-2}
\end{wrapfigure}

\newpage
\clearpage
\begin{wrapfigure}{r}{\textwidth}
  \centering
    \includegraphics[width=0.9\textwidth]{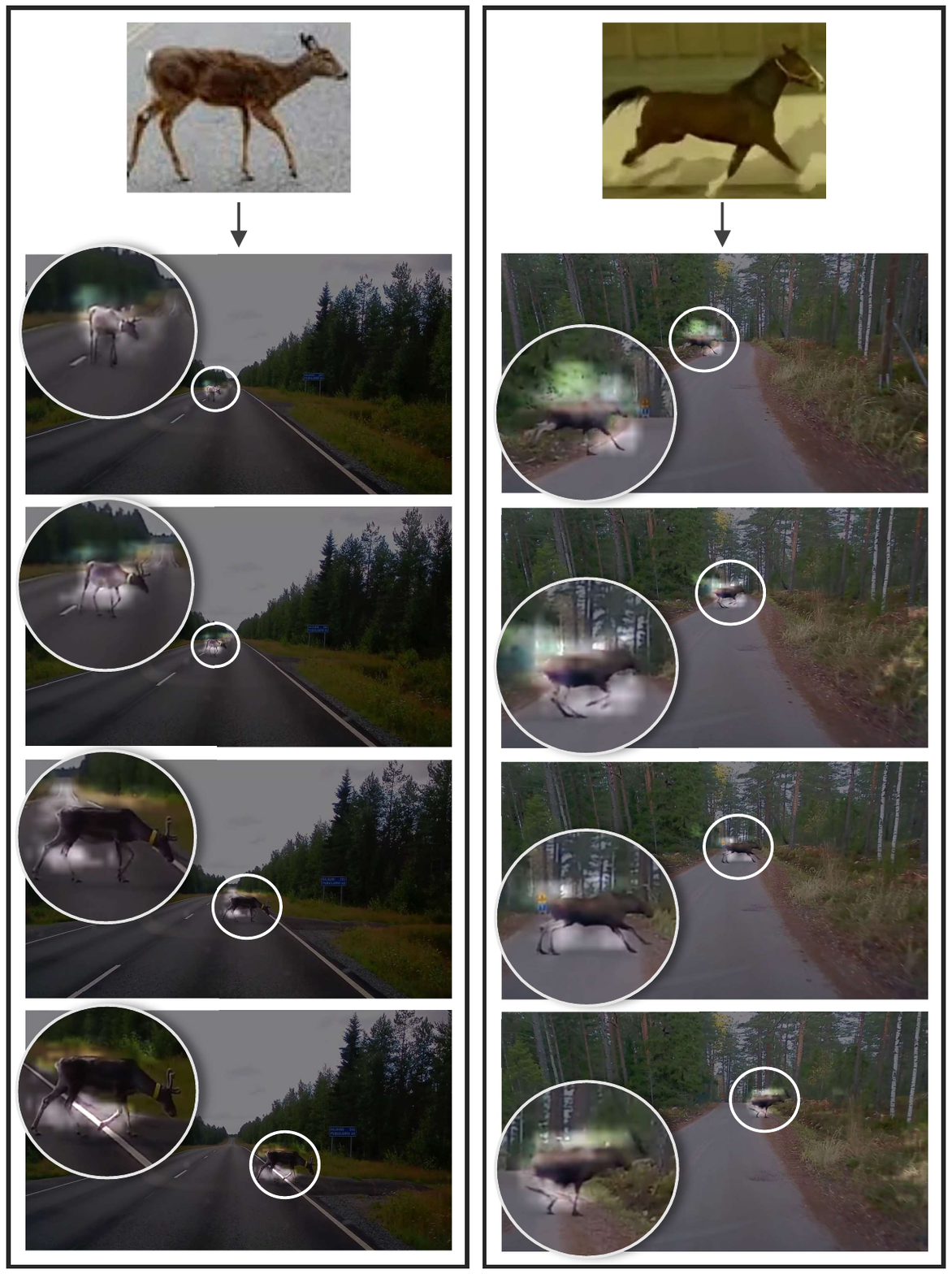}
  \vspace{-0.05in}
  \caption{\textbf{Visualizations on real world scenarios with out-of-domain visual prompts.} At here, we show the similarity map from visual prompt alignment with visual prompts of arbitrary objects unseen during training. Brighter colors highlight higher responses. As illustrated, our method works well in novel scenarios with visual prompts of arbitrary object-of-interests. Best viewed in color.}
  \vspace{-0.05in}
  \label{fig:app-c5-1}
\end{wrapfigure}

\newpage
\clearpage
\begin{wrapfigure}{r}{\textwidth}
  \centering
    \includegraphics[width=0.9\textwidth]{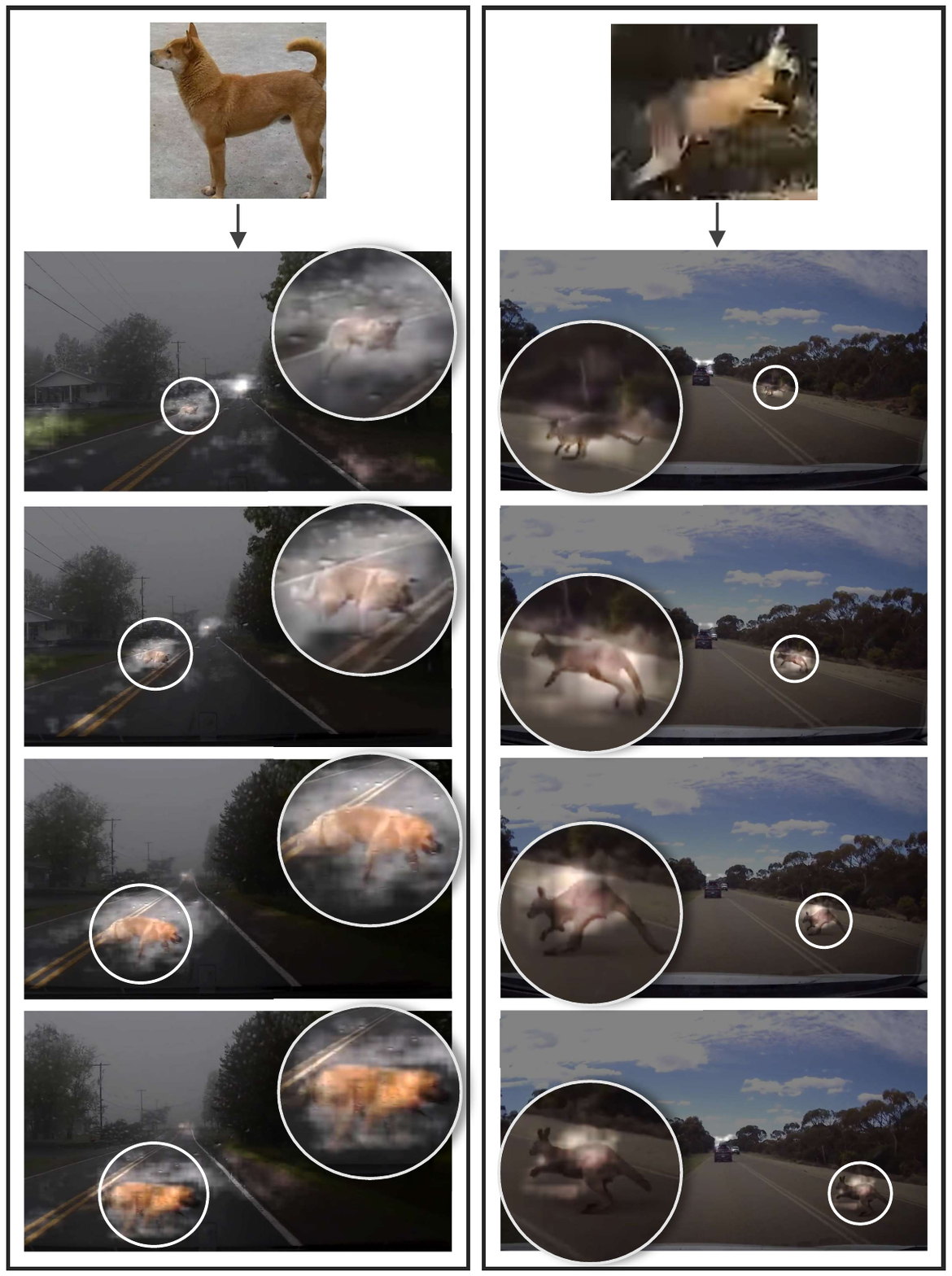}
  \vspace{-0.05in}
    \caption{\textbf{Visualizations on real world scenarios with out-of-domain visual prompts.} At here, we show the similarity map from visual prompt alignment with visual prompts of arbitrary objects unseen during training. Brighter colors highlight higher responses. As illustrated, our method works well in novel scenarios with visual prompts of arbitrary object-of-interests. Best viewed in color.}
  \vspace{-0.05in}
  \label{fig:app-c5-2}
\end{wrapfigure}

\newpage
\clearpage
\begin{wrapfigure}{r}{\textwidth}
  \centering
    \includegraphics[width=0.9\textwidth]{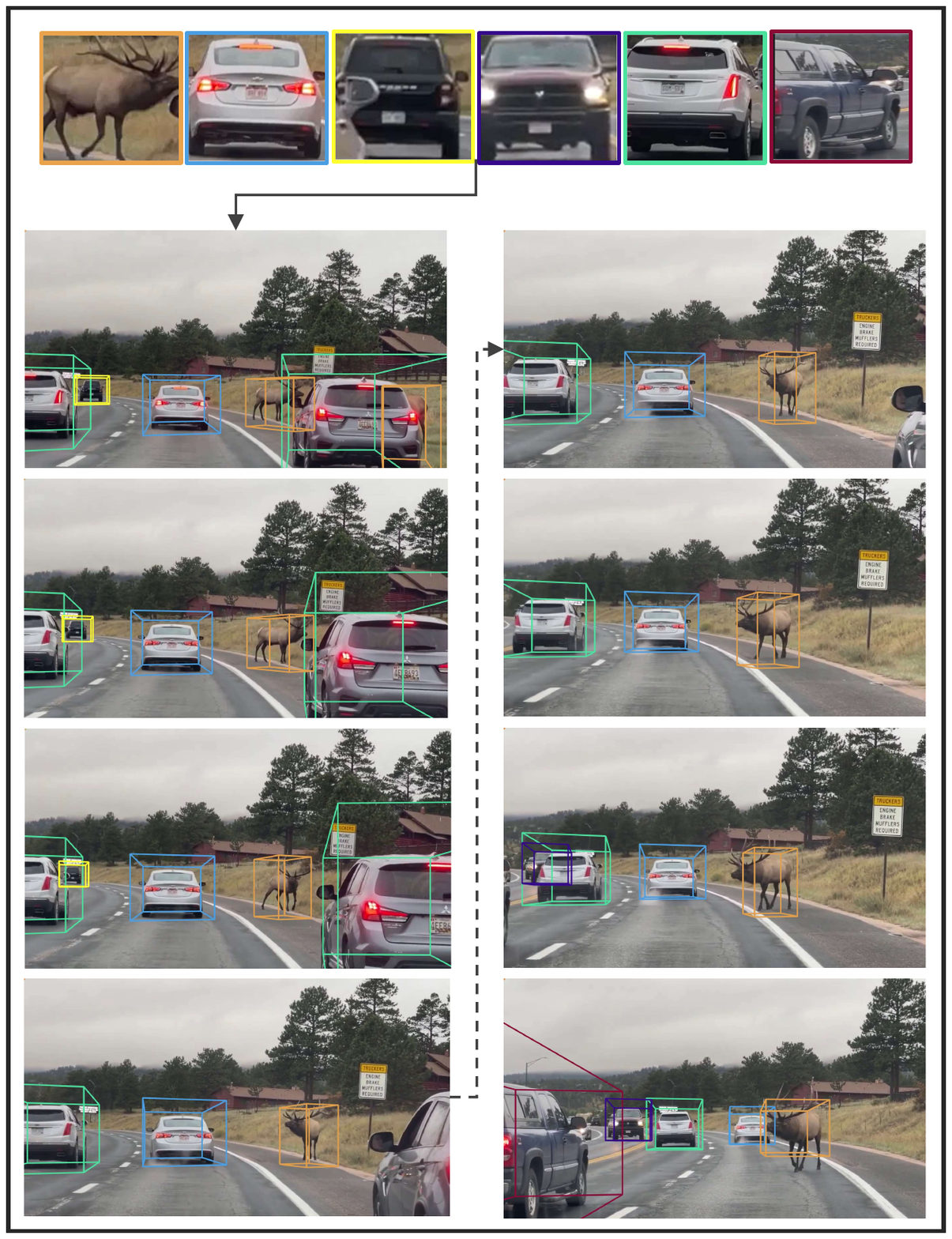}
  \vspace{-0.05in}
    \caption{\textbf{Visualizations on real world scenarios with out-of-domain visual prompts.} This figure demonstrates the \algname's effectiveness in real-world 3D detection, even with visual prompts of unseen objects. demonstrate a case of real-world detection
    As illustrated, our method works well in real-world scenarios with visual prompts of unseen objects. Different object identities are indicated by distinct colors.}
  \vspace{-0.05in}
  \label{fig:app-c5-3}
\end{wrapfigure}

\newpage
\clearpage
\begin{wrapfigure}{r}{\textwidth}
  \centering
    \includegraphics[width=0.9\textwidth]{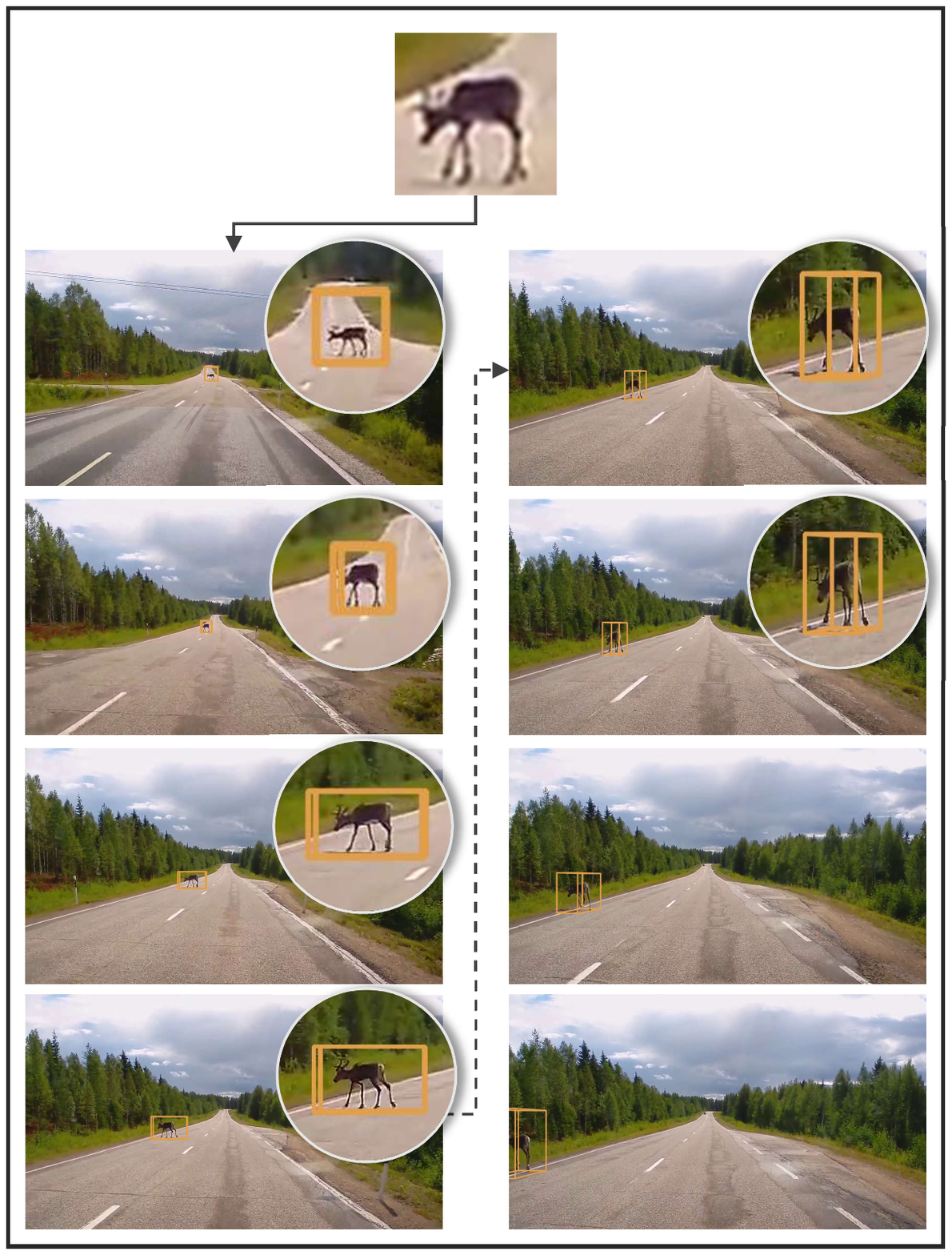}
  \vspace{-0.1in}
    \caption{\textbf{Visualizations on real world scenarios with out-of-domain visual prompts.} 
    This figure demonstrates a case of real-world 3D object detection. We provide the visual prompt from the same scene as the target object, though it is unseen during training, and our \algname system then detects the target objects in the subsequent video frames.
    }
  \vspace{-0.05in}
  \label{fig:app-c5-4}
\end{wrapfigure}

\newpage
\clearpage
\begin{wrapfigure}{r}{\textwidth}
  \centering
    \includegraphics[width=0.9\textwidth]{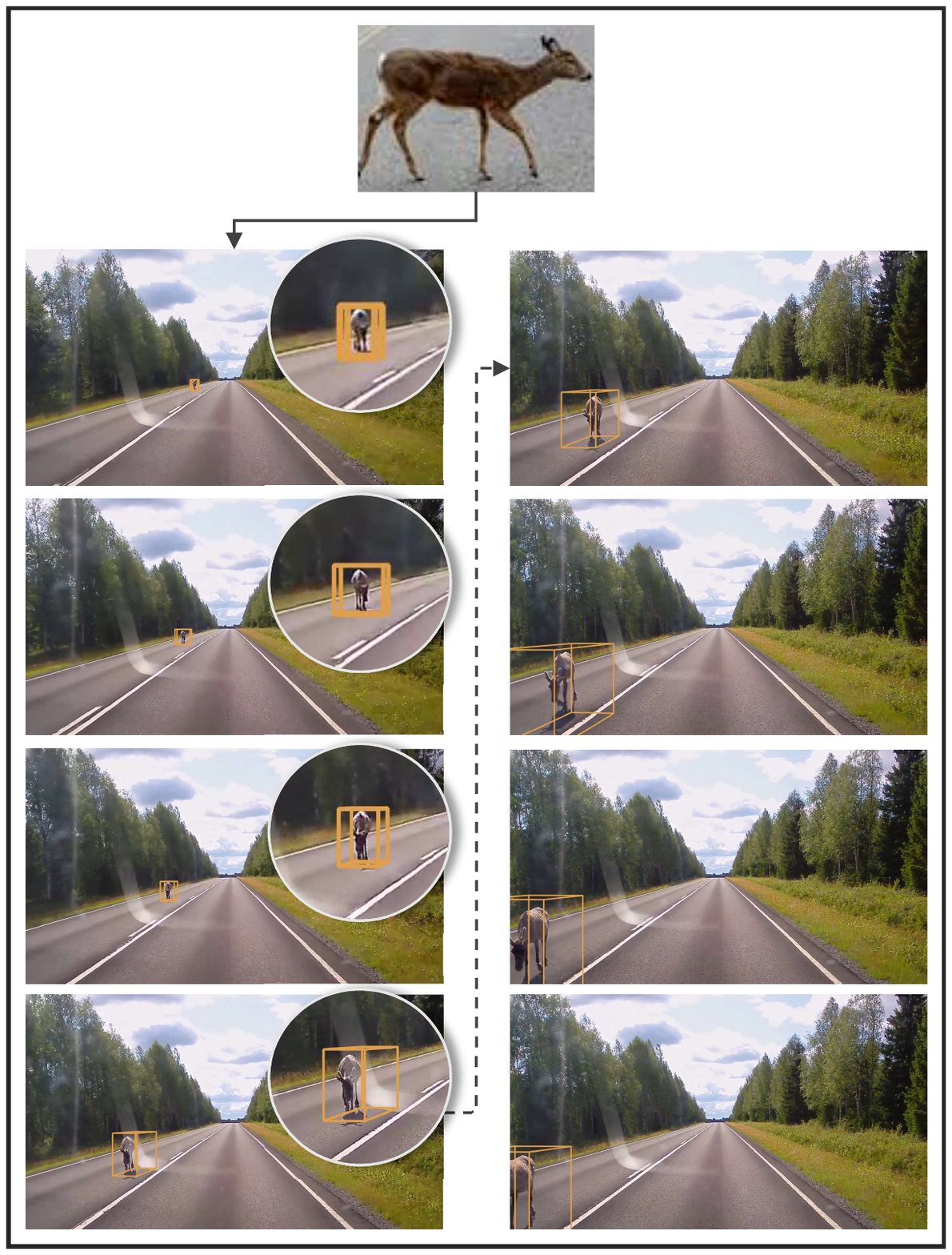}
  \vspace{-0.1in}
    \caption{\textbf{Visualizations on real world scenarios with out-of-domain visual prompts.} This figure demonstrates a case of real-world 3D object detection. We provide a visual prompt from a separate image as the pre-defined prompt and visualize it at the beginning of the sequence.
    As described, our \algname system can successfully detect the "Deer" object using this stylized deer prompt sourced from the internet.
    This example highlights the \algname's capability to effectively leverage diverse, user-supplied visual prompts to accurately identify target objects, even when the prompts are not directly from the same scene.}
  \vspace{-0.05in}
  \label{fig:app-c5-5}
\end{wrapfigure}

\newpage
\clearpage
\begin{wrapfigure}{r}{\textwidth}
  \centering
    \includegraphics[width=0.9\textwidth]{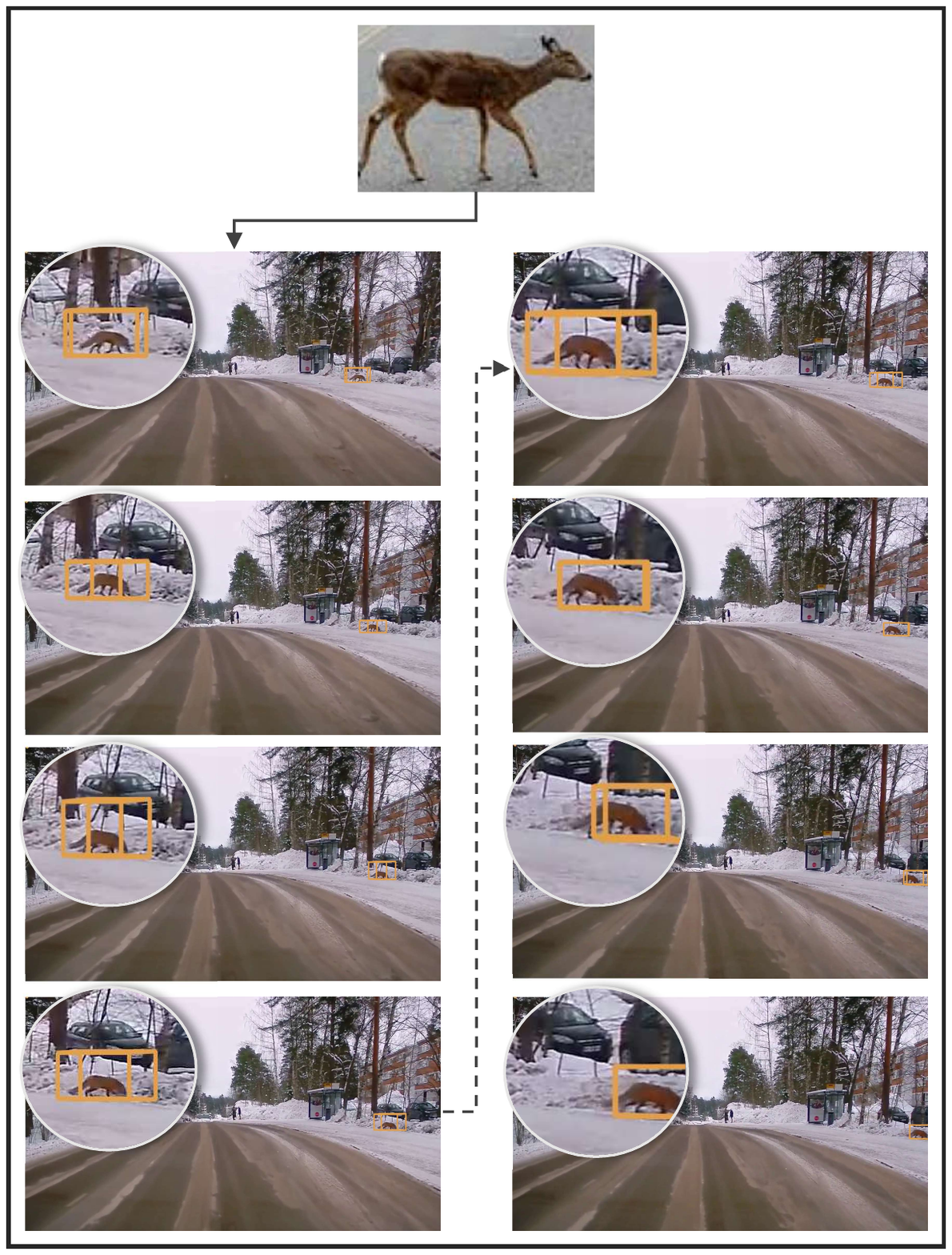}
  \vspace{-0.1in}
    \caption{\textbf{Visualizations on real world scenarios with out-of-domain visual prompts.} 
    Another case demonstrates the use of an internet-sourced "Deer" visual prompt to detect the deer in a different real-world scenario. As shown, our \algname system effectively detects the deer even when it is partially obscured by snow.
    This example further illustrates the robust performance of the \algname framework in accurately localizing target objects, even in challenging environmental conditions, by leveraging flexible visual prompts provided by users.}
  \vspace{-0.05in}
  \label{fig:app-c5-6}
\end{wrapfigure}
}

\end{document}